\documentclass[twoside,11pt]{article}

\usepackage{blindtext}

\usepackage[preprint]{jmlr2e}

\usepackage{comment}
\usepackage{amsmath}
\usepackage{booktabs}
\usepackage{subfigure}
\usepackage{xcolor}
\usepackage{caption}

\usepackage{xcolor}

\definecolor{lightcitationblue}{rgb}{0.2,0.45,0.85}

\hypersetup{
    colorlinks=true,
    citecolor=lightcitationblue,
    linkcolor=black,
    urlcolor=lightcitationblue
}
\newcommand{\FCA}{FewClassArena}
\newcommand{\fcr}{few-class regime}
\newcommand{\added}[1]{#1}

\usepackage{lastpage}
\ShortHeadings{Efficient Neural Network Model Selection for Few-Class Application Datasets}{Efficient Neural Network Model Selection for Few-Class Application Datasets}
\firstpageno{1}

\begin{document}

\title{Efficient Neural Network Model Selection for Few-Class Application Datasets}

\author{\name Bryan Bo Cao \email boccao@cs.stonybrook.edu \\
       \addr Department of Computer Science\\
       Stony Brook University\\
       Stony Brook, NY 11790, USA
       \AND
       \name Abhinav Sharma \email abhinsharma@cs.stonybrook.edu \\
       \addr Department of Computer Science\\
       Stony Brook University\\
       Stony Brook, NY 11790, USA
       \AND
       \name Lawrence O’Gorman \email larry.o\_gorman@nokia-bell-labs.com \\
       \addr Nokia Bell Labs\\
       Murray Hill, NJ 07974, USA
       \AND
        \name Michael Coss \email mike.coss@nokia-bell-labs.com \\
       \addr Nokia Bell Labs\\
       Murray Hill, NJ 07974, USA
       \AND
       \name Shubham Jain \email jain@cs.stonybrook.edu \\
       \addr Department of Computer Science\\
       Stony Brook University\\
       Stony Brook, NY 11790, USA
       }

\editor{My editor}

\maketitle

\begin{abstract}%   <- trailing '%' for backward compatibility of .sty file

% While much effort has been made to develop and benchmark neural network models of high performance, less effort has been made to determine how dataset or data-side properties -- that are known to a practitioner for an application -- can aid in efficient model selection. Neural models are usually benchmarked on datasets of up to 1000s of classes, however many practical applications have fewer than ten classes. To address this understudied but widely practiced segment, complementary to model design, we develop a measure of classification difficulty based upon data-side properties and show how this difficulty metric can be used to identify efficiencies for few-class application datasets that are less attainable than for many-class datasets. We call this ``few-class distinctiveness''. We show how this difficulty metric can be used to compare models and datasets $6$ to $29\times$ faster than through repeated training and testing. Because ``few-class distinctiveness'' reveals better accuracy results than for many-class datasets, we extend scaled model families below the smallest published models to attain more efficiency at similar accuracy levels, for instance 42\% smaller than YOLOv5-nano at similar accuracy for a mobile robot application. Since this work is targeted to small applications with stringent efficiency requirements, we show performance results of few-class model selection for mobile robot, drone, and IoT device examples.

While much effort has focused on developing and benchmarking high-performance neural networks, less attention has been given to how dataset properties, known to practitioners, can guide efficient model selection. Neural models are typically evaluated on datasets with thousands of classes, yet many real-world applications involve fewer than ten. To address this understudied but common setting, we develop a measure of classification difficulty based on data-side properties and show how it enables more efficient model selection for few-class datasets, where traditional approaches are less effective. We term this phenomenon ``few-class distinctiveness''. Our metric allows comparison of models and datasets 6 to $29\times$ faster than repeated training and testing. Leveraging this insight, we extend scaled model families below the smallest published models, achieving greater efficiency at similar accuracy, for example models up to $42\%$ smaller than YOLOv5-nano for a mobile robot task. Targeting resource-constrained applications, we demonstrate few-class model selection across mobile robot, drone, and IoT scenarios, highlighting practical gains in efficiency without sacrificing performance.

\end{abstract}


\section{Introduction}
\label{sec:Introduction}

Part of the rapid progress of neural network technology is due to the public availability of datasets for testing and comparing the performance of different models. Image datasets used for this architecture development usually have many classes (e.g., 1000 for ImageNet \citep{deng2009imagenet}, 80 for COCO \citep{lin2014microsoft}) as is appropriate for general testing. However, these large datasets are often used by practitioners to help select models for their own applications, which may have far fewer classes. In this paper, we shift the focus from the model-side architecture to the data-side or dataset properties, and we show how knowledge of these properties can help choose very efficient models. For one of the data-side properties, number of classes, we offer experimental evidence of a ``distinctiveness'' for classification of few-class application datasets, which we define as those applications with fewer than 10 classes. Knowledge of this distinctiveness can benefit practitioners to select more efficient models than by assuming that many-class dataset performance extrapolates linearly down to few classes.

In the area of computer vision, there are some applications where we want to detect objects from of a dataset of many classes. One example is autonomous driving in which the classes of vehicles, pedestrians, road signs, and road conditions numbers in the 10s or more. However, there are many smaller, practical applications where often two, three, or four is the number of classes in the dataset. The number of classes in a sampling of applications includes six for wildlife animal detection \citep{applWildlifeAustralia2017}; four for cancer cell classification \citep{appl_cancer}; two each for crowd \citep{appl_crowd}, cattle \citep{appl_cattle}, hardhat \citep{appl_hardhat}, and ship \citep{appl_shipSAR} detection; and one for the BioDrone benchmark of single-object tracking for bionic drones \citep{zhao2024biodrone}. In many of these examples, the goal is to design an efficient (energy, memory, time, and cost) \citep{o2025we, kim2025investigating} system with good accuracy. While conventional testing of off-the-shelf models on large datasets can help select between models for accuracy, this will not reliably find the most efficient model without also considering few-class properties.

We identify five independent data-side properties: number of classes, intra-class similarity, inter-class similarity, scale-resolution, and color \citep{cao2023data}. The classification difficulty of an application depends in large part on these properties. For instance, an application whose classes have very similar features will generally have higher classification difficulty than for one with lower inter-class similarity. There are benefits to quantifying this with a difficulty measure based upon dataset properties. Using difficulty metrics associated with datasets, one can compare \textit{just these numbers} instead of performing more time-consuming training and testing trials of many model and dataset combinations. We propose a quantitative, lightweight measure of classification difficulty for an application dataset \citep{cao2024lightweight}, which is based upon the number of classes, intra-, and inter-class similarity.

\begin{comment}
\bo{Strong baseline.}
\bo{``"In our initial effort in this domain, we limit ourselves to serial con- nections and non-adaptive configurations, deferring the investigation, of more flexible model merging to future work." Quoted from the evolutionary merging nature paper.''}
\bo{Comparison with Related Work.}

\bo{Add a concept graph.}
\end{comment}

% \larry{I'm a little confused with this paragraph. Please look at the paragraph I suggest below and choose or modify.}
% \bo{In our initial efforts in this domain, our
% primary contribution lies in characterizing the
% intrinsic comprehensive data-side properties, with
% the goal to assist in lightweight model selection.
% This goal deliberately eschews the development
% of new models or the pursuit of state-of-the-art
% performance in this work. Instead of proposing a
% new model architecture, our work complements
% existing model designs by offering key insights of
% properties from data side that can inspire neural
% architecture innovation, we defer the model design
% to future work.}
The goal in this paper is deliberately \textbf{different from the development of new models} (which is the focus of much important work in the neural network field). Instead, our goal is to take advantage of the \textbf{data-side properties}, known for any application, to choose existing models and those models' parameters that yield efficient, lightweight models most efficiently matched to the application data.
 
% Add this to avoid reviews' mind set of (1) new model design and (2) model architecture comparison.

Contributions of this work are summarized as follows:
\begin{itemize}
\item 
Focusing on data-side properties for efficient model selection that involves establishing a mathematical relationship between classification difficulty and class similarity properties.
\item
Discovery and use of ``few-class distinctiveness'', by which accuracy behaves differently for few-class application datasets than many-class, and which can aid in selecting more efficient models.
\item
Identifying scaled models below those of published model families that offer higher efficiency at similar accuracy for few-class datasets.
\item 
Producing experimental evidence supporting the performance advantages of few-class model selection for efficiency-critical applications of mobile robots, drones, and IoT devices.
\end{itemize}

This paper is organized as follows. In Section \ref{sec:relatedWork}, we discuss related literature on model selection and classification difficulty metrics. In Section \ref{sec:difficMeasure}, a new measure of dataset classification difficulty is determined by maximizing the correlation between class similarity and accuracy across a range of a number of classes. In Section \ref{sec:DataSide}, experiments are performed to understand how number of classes, intra-class similarity, and inter-class similarity affect classification performance. In Section \ref{sec:InPractice}, we describe how sub-models and the difficulty metric are used in practice toward the goal of designing efficient systems, including examples for mobile robot, drone, and IoT (Internet of Things) device applications. And we conclude with limitations and suggested future work in Section \ref{sec:Conclusions}.

%------------------------------------------------------------
\section{Related Work}
\label{sec:relatedWork}

\subsection{Model Selection}
\label{sec:modelSelection}
In recent years, a plethora of neural architectures have been designed, trained, and made easily available such that a practitioner will usually select a model rather than designing or training one from scratch. However, with the increasing number of efficient architectures (e.g. MobileNets \citep{howard2017mobilenets,sandler2018mobilenetv2,koonce2021mobilenetv3}~, SqueezeNet \citep{koonce2021squeezenet}~ ShuffleNets \citep{zhang2018shufflenet,ma2018shufflenet}~, EfficientNet \citep{tan2019efficientnet}, etc.), selecting a proper model to satisfy an application's requirements becomes a challenge in itself. To select a model, existing approaches can be categorized into four options: 1) off-the-shelf, 2) transfer learning plus targeted training, 3) scaled model selection, and 4) selection from model repository. 

For the first option, numerous off-the-shelf models (e.g., \citep{pytorchhub,tfhub,hfhub}) have been trained on datasets containing the same classes as the application, for instance for pedestrian detection \citep{pedestrian2017}, flower classification \citep{flowers2018}, fish classification \citep{fish2018}, and food detection \citep{food2018}. Although this can save substantial time from data collection, labeling, and training, it often fails in real-world applications because of feature shift \citep{huseljic2024systematic} in deployed environments caused by such factors as different camera capture angles, backgrounds, object scales, etc.

A second option is transfer learning \citep{transferLearn2021, zhao2025efficient} by which a model is trained on a larger, standard dataset such as ImageNet \citep{deng2009imagenet} or COCO \citep{lin2014microsoft}, stripped of its classification layer (leaving the backbone), then fine-tuned \citep{ngnawe2025robust} on objects of the application of interest. An efficiency drawback to this popular choice is that the backbone incorporates extraneous features for the large dataset that are not needed for the application classes. Although transfer learning facilitates the selection of a neural model with high accuracies, the model will be larger than one trained on only the classes of interest for the same level of accuracy.

A third option is to choose from a scaled model family -- a collection of models that share the same general architecture, but whose size (width, depth, and resolution) are adjusted with a scaling factor. Examples include the EfficientNet family \citep{tan2019efficientnet}~ whose scale levels range from B0 (small) to B7 (large), the YOLO family \citep{glenn_jocher_2021_5563715} whose scale levels range from ``nano'' to ``extra-large'', and the LPS-Net family \citep{zhang2023lightweight} whose scale levels range from ``S'' to ``L''. Since this paper focuses on efficiency for small, few-class applications, we experiment and compare on the smaller levels of the EfficientNet image classifier (EfficientNet-B0) and the YOLO object detector (YOLO-nano) \citep{ganesh2022yolo}.

The fourth option is to select a pre-trained model from a model repository \citep{SommelierModelRepository_2022,zhou2023learnware, huseljic2024systematic}. This is a fast way to start using a model, but it is different than the focus of this paper in two ways. Our focus is on selecting efficient models to match dataset properties. The most efficient models must be trained on the dataset of the application. This is unlike a  model repository whose models may be efficient, but are unlikely to be the \textit{most efficient} (accuracy versus model size) for the \textit{specific application dataset}. Contrary to class-incremental learning (CIL) \citep{zhou2024revisiting}, but in line with the spirit of semantic image matting \citep{sun2024semantic}, we aim to retain only the deployed dataset-specific semantics from a general one. The second difference is that our methods are directed to the \textit{data side} versus the \textit{model side} \citep{ZestForLIME_2022,modelDiff2023}. One benefit is that this focus enables different application datasets to be quickly compared by their classification difficulty metrics rather than only by their test performance after training on that dataset, the latter being much more time-consuming.

\subsection{Image Classification Difficulty}
\label{sec:imClassBG}
No matter which type of classifier is used, their empirically-determined behavior is strongly data dependent. Previous to the widespread adoption of neural networks, classification difficulty was largely measured by the ability to distinguish classes volumetrically in multi-dimensional feature space. A classical measure is Fisher's Discriminant Ratio, by which a large difference in class means (inter-class) and a small sum of their variances (intra-class) describes a less difficult classification problem \citep{dudaHart1973}. In \citep{ClassComplexityHo2002}, complexity measures are described that include feature overlap, separability of classes, and geometry of the class manifolds. Although the embedding space of a neural network is also a feature space, neural networks often have much higher dimensionality of nonintuitive (machine-learned) features, which have the ability to better distinguish highly non-linear class boundaries, thus leading to neural network classification difficulty metric different from these previous metrics.
 
A number of image difficulty measures have been proposed. For metric learning \citep{bellet2013survey,metricLearn2017,SimilarityMetricLearning2021}, the loss function is set to minimize the similarity between images of the same class during training, where similarity is measured as the dot product between embeddings of two images (usually from the last hidden layer of the model) \citep{Ionescu_2016_CVPR,medDiagnose_2022,AngularGap2022}. Another way to measure difficulty is by classification error on a difficulty-scaled range of datasets \citep{DatasetClassification_2021,imagenetDifficulty2022,DLLagree2022} or models \citep{ScaledInference_2022}. Machine difficulty scores can also be used to prune filters associated with easier features during training \citep{dynamicPrune_2023}, and by more highly weighted filters of difficult features during inference \citep{weightedModel2020}.

One difference from these previous papers is that we focus upon \textit{classification difficulty}. Many references calculate \textit{single-image} difficulty for purposes of curriculum (or simple-to-difficult) learning \citep{Ionescu_2016_CVPR,appalaraju2017image,medDiagnose_2022,AngularGap2022} and scaled model selection \citep{Ionescu_2Stage_2018}. In \citep{weightedModel2020}, intra-class difficulty is measured for the purpose of weighting classes differently during training. In contrast to single-image difficulty, we incorporate intra- and inter-class similarity in determining a difficulty metric for application datasets containing many images of multiple classes, as shown in Section \ref{sec:difficMeasure}. And in Section \ref{sec:DataSide}, we show by experiment how the measure varies for different numbers of classes and similarities.

\subsection{Relation to Representation Learning Frameworks}

\noindent \textbf{\added{Few-Shot Learning.}}
The goal of the Few-Short Learning (FSL) framework \citep{song2023comprehensive, wang2020generalizing, hu2022pushing, sung2018learning, wang2024clip, wu2024few, lang2024few} is to leverage the representations learned from very few samples or prior knowledge for generalizing to other tasks or domains.

\noindent \textbf{\added{Self-Supervised Learning.}} 
Under the Self-Supervised Learning (SSL) paradigm, tasks such as pretext mask-and-reconstruct methods \citep{he2022masked} and Contrastive Learning \citep{chen2020simple, he2020momentum, chen2020improved, chen2021empirical, gao2024clip, zhu2024weakclip} are designed to generate supervisory signals from the dataset itself, enabling the learning of general visual features \citep{jaiswal2020survey, jing2020self}.

We distinguish our work from the aforementioned learning frameworks by focusing on target application deployment. Specifically, we aim to identify the most \textbf{efficient} baseline model that achieves the task using the \textbf{minimal} set of required features.

% In contrast to the advancements of the aforementioned learning frameworks, this work focuses on the research problem of selecting the most \textbf{efficient} model with \textbf{minimal} features (e.g. model parameters, model multiplies) needed for the target application deployment.}

% From ICLR25 >>
\begin{comment}
\noindent \textbf{\added{Few-Shot Learning.}}
\added{There has been a large body of research on Few-Short Learning (FSL) \citep{song2023comprehensive, wang2020generalizing, hu2022pushing, sung2018learning}. However, the fundamental research questions differ from ours in the \fcr~. The FSL framework aims to address the problem of \textbf{data scarsity} with the goal for a model to leverage the representations from very few samples (or none, in the case of Zero-Shot Learning), or prior knowledge that can \textbf{generalize} effectively to other tasks or domains.}

\noindent \textbf{\added{Self-Supervised Learning.}} \added{To leverage the knowledge from unlabled data, Self-Supervised Learning (SSL) has emerged as an effective learning framework to learn general vision features \citep{jaiswal2020survey, jing2020self}. This includes techniques such as Contrastive Learning, applied to single modalities \citep{chen2020simple, chen2021exploring, he2020momentum, chen2020improved, chen2021empirical} or multiple modalities \citep{radford2021learning}, as well as mask-and-reconstruct methods \citep{he2022masked}, among others.}
\end{comment}
% From ICLR25 <<

%--------------------------------------------------

\section{Classification Difficulty Metric}
\label{sec:difficMeasure}

We first define a classification difficulty metric. This will be used in subsequent sections to support three contributions of this paper: i) few-class distinctiveness, ii) correlation between class similarity and accuracy, and iii) a metric for quickly comparing models.

Cosine similarity is a common measure used to determine the closeness between two feature vectors, 
\begin{equation}
%\footnotesize
    \textrm{s}(\textbf{x}, \textbf{y}) = 
    \cos(\theta_{xy}) = {\textbf{x}}^T {\textbf{y}} / 
    ( ||{\textbf{x}}|| \: ||{\textbf{y}}|| ).
     \label{eqn:cosineSimilarity}
\end{equation}

Cosine similarity has a range of $[-1.0, 1.0]$, however in practice, most cosine similarity values between two natural images are positive. Normalization to $[0.0, 1.0]$ can be accomplished by a $softmax$ function as in \citep{medDiagnose_2022}, but that reduces the positive range by $1/e$, so we use a $ReLu$ function, 
\begin{equation}
%\footnotesize
    s'{(\textbf{x}, \textbf{y})}
    = \textrm{ReLu} \bigl( s{(\textbf{x},  \textbf{y})} \bigr) \\
    = \textrm{max} \bigl[ s{(\textbf{x},  \textbf{y})}, 0 \bigr].
     \label{eqn:relu}
\end{equation}

For the purpose of defining a classification difficulty metric, we distinguish intra- and inter-class similarity, the first that relates inversely to difficulty and the latter that relates directly to difficulty. We define \textit{intra-class similarity} as the average of similarities between all pairs of instances in the same class,
\begin{equation}
%\footnotesize
    S_R(\{\textbf{x}\}) = \frac{1}{n_1}\sum_{i,j \in C_X, i \neq j }s'(\textbf{x}_{i}, \textbf{x}_{j}),
     \label{eqn:SR}
\end{equation}
where $C_X\{\textbf{x}\}$ is a set of instances in class $C_X$, and $n_1$ is the number of pair combinations of instances in that set.
We define \textit{inter-class similarity} as the average of similarities of all pairs of instances between two classes,
\begin{equation}
%\footnotesize
    S_E(\{\textbf{x}\} \{\textbf{y}\}) = \frac{1}{n_2}\sum_{i \in C_X, j \in C_Y}s(\textbf{x}_{i}, \textbf{y}_{j}),
     \label{eqn:SE}
\end{equation}
where $C_Y\{\textbf{y}\}$ is a set of instances in class $C_Y$, and $n_2$ is the number of pair combinations of instances between sets $C_X$ and $C_Y$. 

% MAX ----------------
\iffalse
For inter-class similarity, instead of averaging as in equation \ref{eqn:S2}, some references \citep{rousseeuw1987silhouettes,AngularGap2022} use the maximum similarity between the $n_2$ instance pairs as in, \larry{This is wrong. It's max class similarity.}
\begin{equation}
%\footnotesize
    S_{EM}(\{\textbf{x}\} \{\textbf{y}\}) = \max\limits_{i \in C_X, j \in C_Y}s(\textbf{x}_{i}, \textbf{y}_{j}),
     \label{eqn:SEM}
\end{equation}
\fi

Equations \ref{eqn:SR} and \ref{eqn:SE} are for one and two classes. For the general case of $N_{CL}$ classes (and assuming balanced set sizes), we can take averages of those equations to obtain the average intra- and inter-class similarities:
%, and inter- (with maximum) similarities:
\begin{equation}
%\footnotesize
    \textbf{Intra-class: }\bar{S}_R = \frac{1}{N_{CL}}\sum_{i} S_{Ri}(\{\textbf{x})\},
     \label{eqn:S1Avg}
\end{equation}
\begin{equation}
%\footnotesize
    \textbf{Inter-class: }\bar{S}_E = \frac{1}{n_3}\sum_{j}S_{Ej}(\{\textbf{x}\} \{\textbf{y}\}), % \hspace{0.25cm}
    % \bar{S}_{2M} = \frac{1}{n_3}\sum_{j}S_{2Mj}(\{\textbf{x}\} \{\textbf{y}\}),
     \label{eqn:S2Avg}
\end{equation}
% MAX ----------------
\iffalse
\begin{equation}
%\footnotesize
    \textbf{Inter-class (max): }\bar{S}_{EM} = \frac{1}{n_3}\sum_{j}S_{EMj}(\{\textbf{x}\} \{\textbf{y}\}),
     \label{eqn:SEM}
\end{equation}
\fi
where $0 \leq i < N_{CL}$, $0 \leq j < n_3$, and $n_3 = \genfrac(){0pt}{1}{N_{CL}}{2}$ is the number of combinations of class pairs.

Variations of difficulty measures can be found in the literature for purposes that are similar, but not exactly the same as classification difficulty (as discussed in Section \ref{sec:imClassBG}). Below, we follow a progression through five of these, and compare them with our new proposal. All the difficulty measures $\mathcal{D}_i$ are functions of the average intra- and/or inter-class similarities. We correlate the measures with experimental accuracy results to determine which measure is best to predict classification accuracy. All difficulty measures have range $[0,1]$. 

For completeness, we begin with difficulty metrics based upon only one of inter- and intra-class similarity each. Since inter-class similarity is directly related to classification difficulty, the first metric is,
\begin{equation}
%\footnotesize
    \mathcal{D}_A 
    = \bar{S}_{E}.
     \label{eqn:DifficA}
\end{equation}
And since intra-class similarity is related as the complement to difficulty, the second metric is,
\begin{equation}
%\footnotesize
    \mathcal{D}_B 
    = 1.0 - \bar{S}_{R}.
     \label{eqn:DifficB}
\end{equation}

Both similarity metrics can be combined into a proportional relationship as in reference \citep{medDiagnose_2022}. For $D_C$, $\bar{S}_E$ is normalized by both similarities,
\begin{equation}
%\footnotesize
    \mathcal{D}_C 
    = \bar{S}_{E} / (\bar{S}_E + \bar{S}_R).
     \label{eqn:DifficC}
\end{equation}
Similarly for $D_D$, 
\begin{equation}
%\footnotesize
    \mathcal{D}_D 
    = (1.0 - \bar{S}_{R}) / (\bar{S}_E + \bar{S}_R).
     \label{eqn:DifficD}
\end{equation}

Instead of a proportional relationship of similarities as in equations \ref{eqn:DifficB} to \ref{eqn:DifficD}, \citep{AngularGap2022} defines difficulty with a difference relationship between similarities,  
\begin{equation}
%\footnotesize
    \mathcal{D}_{E} = 
    0.5 \bigl (1.0 - \bar{S}_R 
    + \bar{S}_{E} \bigr).
     \label{eqn:DifficE}
\end{equation}
% MAX------------
\iffalse
Using the maximum of inter-class similarity, \larry{Figure out if max is worthwhile keeping.}
\begin{equation}
%\footnotesize
    \mathcal{D}_{6} = 
    0.5 \bigl (1.0 - \bar{S}_1 
    + \bar{S}_{2M} \bigr),
     \label{eqn:Diffic6}
\end{equation}
\fi

\begin{figure}[t]
\centering
    \includegraphics[width=1.0\linewidth]{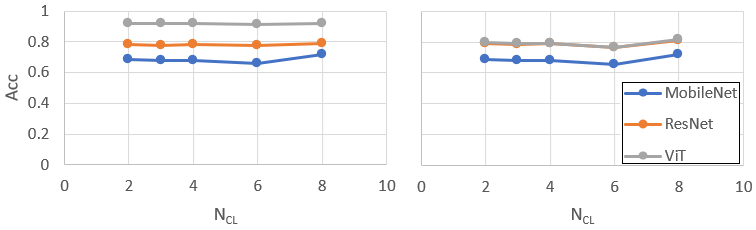}
   \caption{Accuracy results over a number of classes from 2 to 8 and for three models. Results for the CIFAR-100 dataset is on the left, and for ImageNet-200 on the right.}
   \label{fig:accuracies}
\end{figure}

To determine how each difficulty metric relates to classification accuracy, we performed the following experiments. We chose three neural network classifier models that range in size: MobileNet-V2 (small) \citep{sandler2018mobilenetv2}, ResNet-50 (medium) \citep{he2016deep}, and ViT (large) \citep{hfvit2020}. We chose two datasets: CIFAR-100 and ImageNet-200. Each model was trained and tested on subsets of the datasets that comprised 2, 3, 4, 6, and 8 classes. We grouped 200 randomly chosen subsets for each dataset and for each number of classes. We show the accuracy results in Fig. \ref{fig:accuracies} to show that the results span a range to be expected for the models and datasets. These show: i) accuracy is directly related to model size (ViT better than ResNet-50 better than MobileNet), and ii) the larger dataset, ImageNet-200, is more challenging than the smaller dataset, CIFAR-100.

\begin{figure}[t]
\centering
    \includegraphics[width=1.0\linewidth]{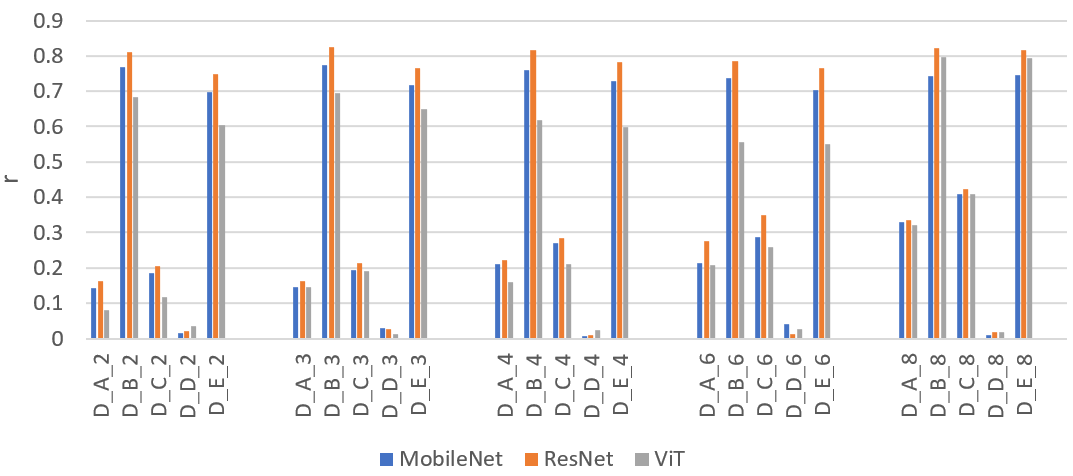}
   \caption{Spearman correlation values on the CIFAR-100 dataset for three models and for class sizes $N_{CL}=2,3,4,6,8$. On the horizontal axis, ``D\_A\_2'' designates ``Difficulty method A, 2 classes''; ``D\_B\_3'' designates ``Difficulty method B, 3 classes''; etc.}
   \label{fig:R_bar_DvsAccCIFAR}
\end{figure}

\begin{figure}[t]
\centering
    \includegraphics[width=1.0\linewidth]{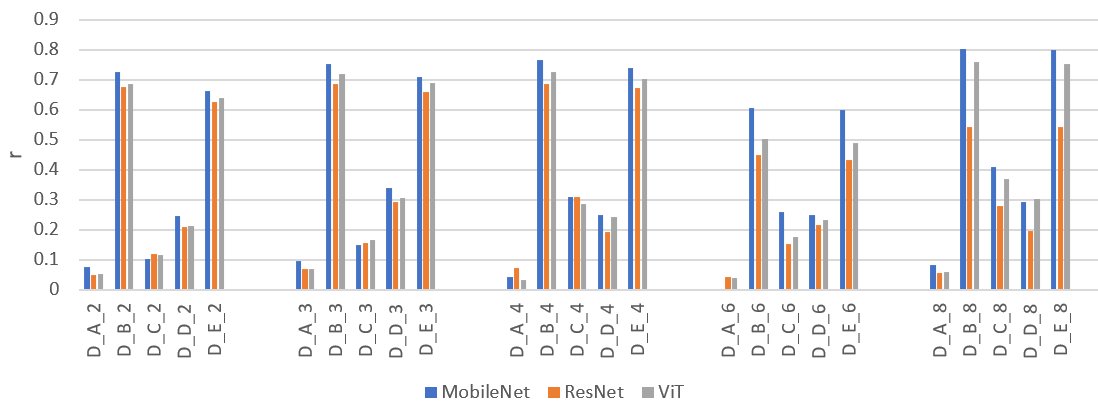}
   \caption{Spearman correlation coefficients on the ImageNet-200 dataset for three models and for class sizes $N_{CL}=2,3,4,6,8$. Horizontal axis designations are as in Fig. \ref{fig:R_bar_DvsAccCIFAR}.}
   \label{fig:R_bar_DvsAccImageNet}
\end{figure}

The five difficulty values are calculated for each model/dataset/subset/tuple size, so there are 3 models x 2 datasets x 200 subsets x 5 tuple sizes (total 6000) pairs of accuracy results and difficulty values. We perform a Spearman correlation for these pairs. Spearman correlation measures the correlation of ranking between values, not the values themselves. So for our purposes it is used to predict whether one dataset/model combination will have better accuracy \textit{relative} to another. Results are shown for the CIFAR-100 dataset in Fig. \ref{fig:R_bar_DvsAccCIFAR} and the ImageNet-200 dataset in Fig. \ref{fig:R_bar_DvsAccImageNet}.

In Fig.s \ref{fig:R_bar_DvsAccCIFAR} and \ref{fig:R_bar_DvsAccImageNet}, it can be seen that methods $D_B$ and $D_E$ have much higher correlation coefficients than the other methods. (A value above 0.7 is considered to indicate strong correlation \citep{schober2018correlation}.) We can do a further experiment to more precisely determine if and how the proportions of intra- and inter-similarity affect correlation with accuracy by replacing the 0.5 multiplier in equation \ref{eqn:DifficE} with complementary weights $\gamma$ and $1.0 - \gamma$, where $0 \le \gamma \le 1.0$, on intra- and inter-class similarity, 
\begin{equation}
%\footnotesize
    \mathcal{D}_{F} = 
    \gamma (1.0 - \bar{S}_R) + (1.0 - \gamma) \bar{S}_{E}.
     \label{eqn:DifficF}
\end{equation}

\begin{figure}[t]
\centering
    \includegraphics[width=1.0\linewidth]{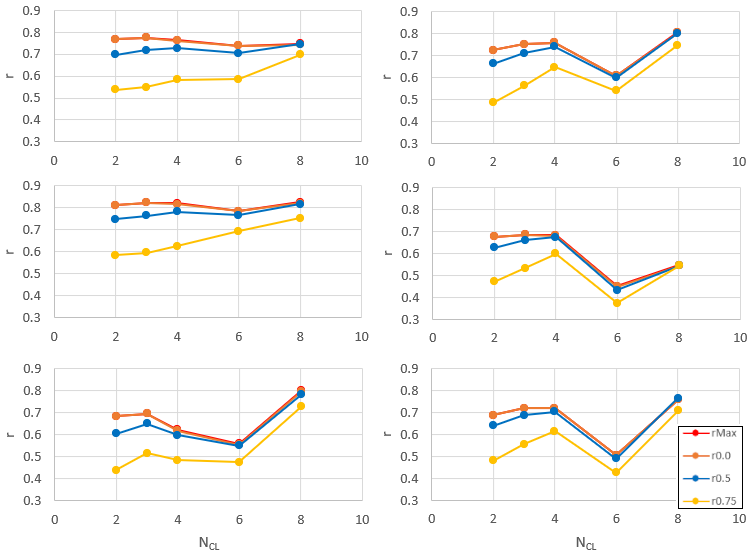}
   \caption{Spearman correlation coefficient values as a function of the number of classes, $N_{CL}$. The left column of plots are for the CIFAR-100 dataset, and the right are for ImageNet-200. The top row of plots is for the MobileNet-V2 model, middle is for ResNet-50, and bottom is for ViT. The legend indicates weight values for the plots of $\gamma = \{0.0, 0.5, 0.75\}$, and $r_{max}$ indicates the best-case correlation with respect to variable weight values.}
   \label{fig:RVsNc6plots}
\end{figure}

We performed experiments to see how the weight $\gamma$ in equation \ref{eqn:DifficF} affects the correlation between difficulty and accuracy across the range of few classes. Figure \ref{fig:RVsNc6plots} displays the results of Spearman correlation versus the number of classes for $\gamma = \{0.0, 0.25, 0.5, 0.75\}$ as well as for the weight where correlation is maximum. As before, we tested all combinations of the models MobileNet-V2, ResNet-50, and ViT, with the CIFAR-100 and ImageNet-200 datasets. We tested for the few-class size range of $N_{CL} = \{2, 3, 4, 6, 8\}$.

Two observations can be made from the results in Fig. \ref{fig:RVsNc6plots} that are consistent among the three models and two datasets. The first is that the highest correlation values are for the weight range within $0.0 \le \gamma \le 0.5$. Consistently, all the plots for $\gamma =0.75$ have lower correlation than $0.0 < \gamma < 0.5$. The second is that the correlation distance between plots narrows as $N_{CL}$ goes from 2 to 8. This means that the weight value in the difficulty equation matters less for measuring correlation at higher $N_{CL}$ than lower $N_{CL}$ within the range tested. We use these observations in Section \ref{sec:DataSide} to describe properties and in Section \ref{sec:InPractice} to describe methods associated with the few-class applications.

%--------------------------------------------------

\section{Data-Side Properties}
\label{sec:DataSide}

To reiterate a main message of this paper, the practitioner can take advantage of knowledge of data-side properties to help select efficient models. In this section, we describe the relationships between data-side properties and system performance. The data-side properties are: number of classes, intra-class similarity, inter-class similarity, color, and ``scale-resolution''. Note that while the two similarity properties are independent, scale and resolution are inter-dependent, so we use the term ``scale-resolution'' for this single property.

%-------------------------------------------------%
\subsection{Number of Classes}
\label{sec:nCl}

Since the focus of this paper is few-class datasets, our first two tasks are the following, i) to show that few-class datasets have properties distinct from datasets of more classes, and ii) to define the boundary between ``few'' and ``many'' classes.

We performed the following experiment. We chose the ResNet model because it is widely used and has a range of sizes that have been shown to directly relate to performance across many datasets \citep{he2016deep}. We chose the ImageNet-1000 dataset because it is also widely used and its 1000 classes enables us to test subsets of the full dataset from 1000 down to the 2 classes. We performed two sets of tests. For a baseline, we trained the full dataset and tested on subsets of classes with this same model. We call this \textit{full-dataset training}. Compared with this, we trained subsets for a range of numbers of classes. We call this \textit{subset-only training}. For each number of classes, we trained upon 5 random selections of classes and averaged the accuracy results. In total, we trained and tested 5 ResNet models from ResNet-152 to ResNet-18, 10 sizes of number of classes, and 5 subset selections per $N_{CL}$, a total of 250 models.
% https://docs.google.com/spreadsheets/d/1ZqYd92BB7ljOAoDSz96623dwF-9LqV3UzWX2YHNf9Zs/edit?gid=2093056462#gid=2093056462
% bryan.bo.cao.s@gmail.com
% ResNet Models: ResNet18, 34, 50, 101, 152
% 5 subsets: seed #0-4
% #Cls: 2,3,4,5,10,100,200,400,600,800

\begin{figure}[h]
\centering
    \subfigure[Blue plot (subset-only training) increases in accuracy for fewer classes, whereas red plot (full dataset training) decreases.]
    {\includegraphics[width=0.6\linewidth]{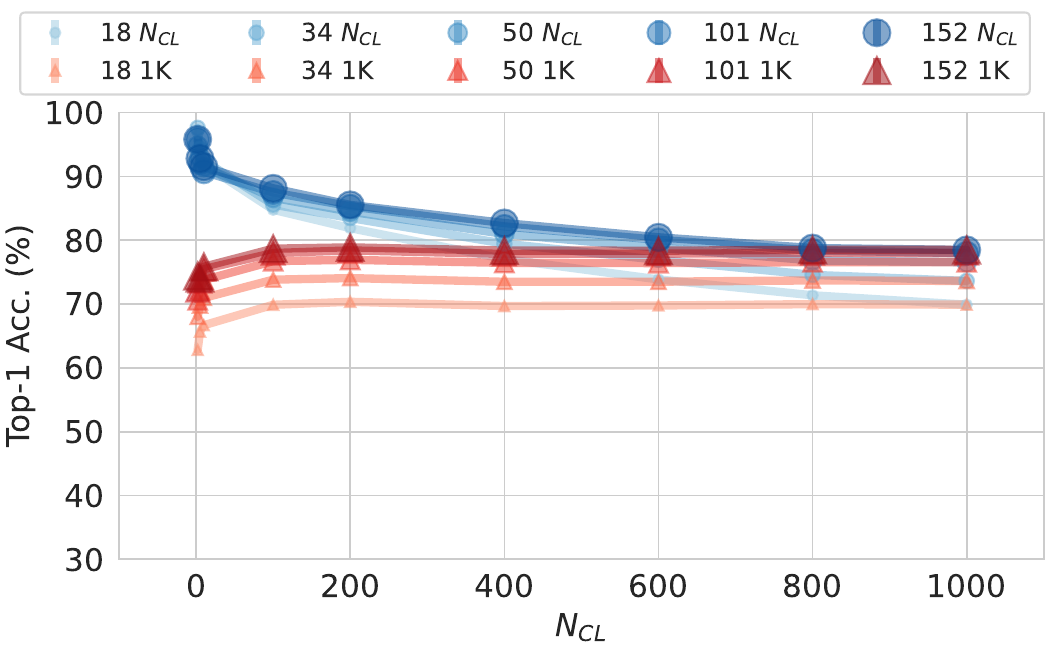}}
    \subfigure[Results for subset-only training. Inset shows change in the increase in accuracy for fewer than 10 classes.]
    {\includegraphics[width=0.6\linewidth]{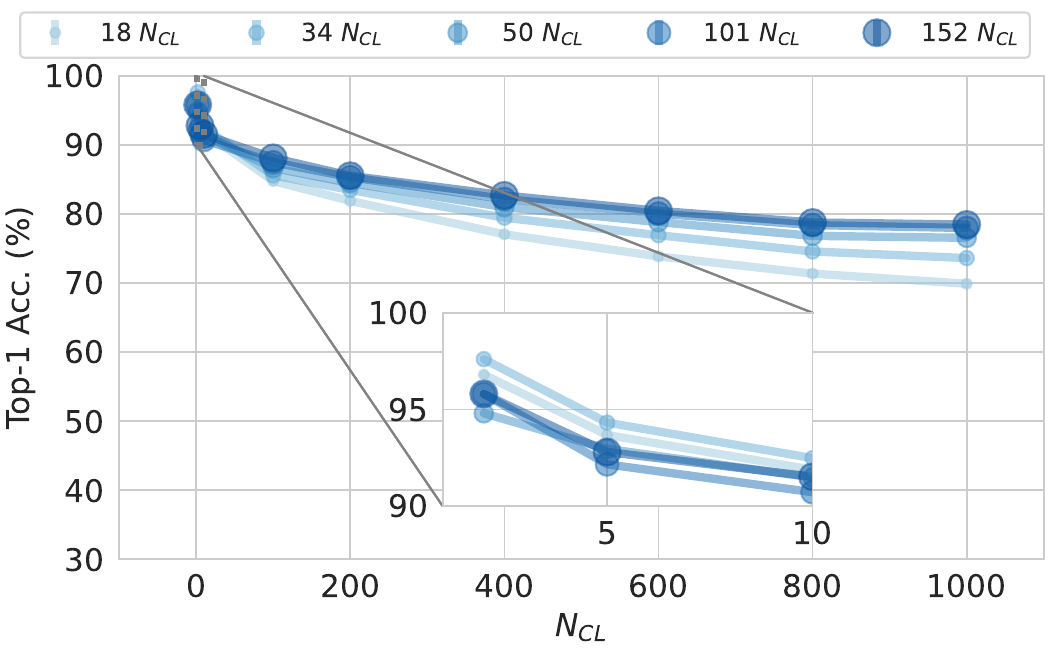}}
    \subfigure[Results for full-dataset training. Inset shows 
 change in the drop of accuracy for fewer than 10 classes.]
    {\includegraphics[width=0.6\linewidth]{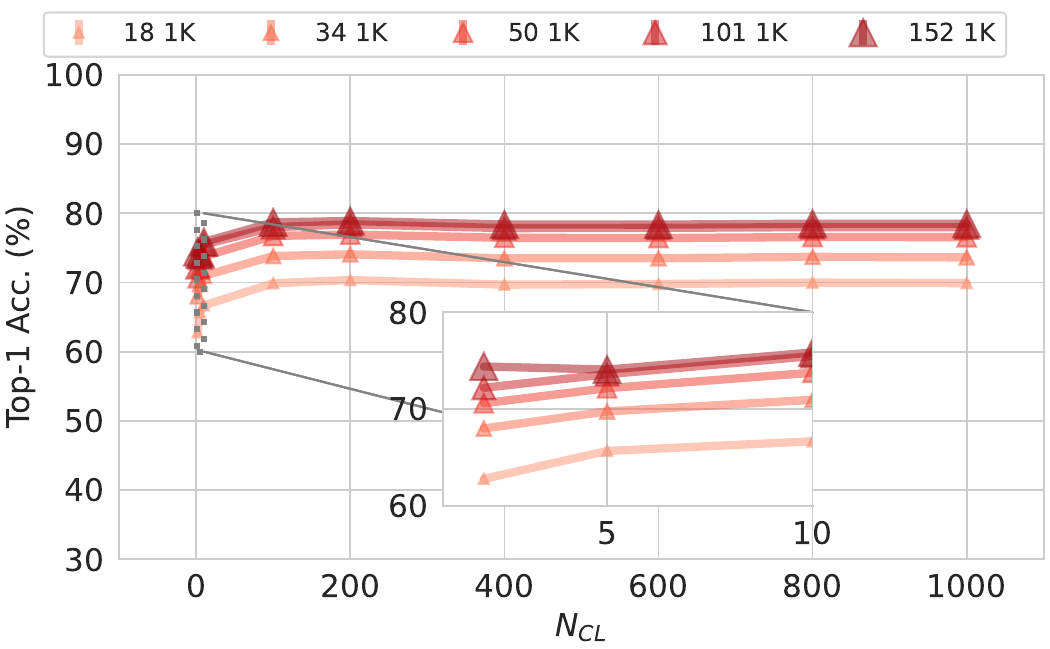}}
   \caption{Top-1 accuracy results over a range of number of classes for ResNet on the ImageNet dataset. The red plots are for full-dataset training and the blue for subset-only training. The legend above each graph shows the range of ResNet model sizes tested, and these are plotted with saturation (of red and blue) decreasing as model size decreases.}
   \label{fig:ResNet_fewClass}
\end{figure}

Results of the ResNet tests in Fig. \ref{fig:ResNet_fewClass} can be used to address the two tasks of the section. i) Few-class datasets have different properties than for more classes. Evidence in the plots of Fig. \ref{fig:ResNet_fewClass} shows that accuracy \textit{improves} for few classes with subset-only training (blue) as class number decreases, whereas accuracy \textit{decreases} in this range for full-dataset training (red). There is lower variance in accuracy results among the five ResNet models for subset-only training (the inset of the middle graph shows about 3\% span among model results), whereas there is higher variance for full-dataset training (the inset of the bottom graph shows about 10\% span among model results). Regarding the second task, ii) we designate the region below $N_{CL} = 10$ as the ``few-class regime'', a concept closely related to, but distinct from ``narrow domains'' \citep{michaud2025creation}. 
% Although this choice is somewhat arbitrary seeing as the plot has most extreme difference in rise for the very few number of classes, 2 to 4, 
We use a number of 2 to 10 classes to perform more complete examination on this wider range as we do in our experiments for completeness in this paper.

We can make further observations from Fig. \ref{fig:ResNet_fewClass}. i) There is a steady improvement of accuracy from $N_{CL} = 1000$ down to the few-class regime for the subset-only training curve, whereas the full-dataset training curve stays relatively flat from $N_{CL} = 1000$ down to the few-class regime. ii) Both subset-only training and full-dataset training curves follow the scaling law \citep{neuralScaling2024} for $N_{CL} = 1000$ down to -- but not including -- the few-class regime, where larger models (dark blue or red) have higher accuracies than smaller models (lighter blue or red). Within the few-class regime, the full-dataset training curves adhere to this scaling law, but the subset-only training curves do not (see the middle plot). This may be due to the much lower deviation in accuracy values between model sizes for few classes.

\begin{figure*}[ht]
  \begin{minipage}{1\linewidth}
  \centering

    \subfigure[Classification on CIFAR-100 with two ResNets.]        %%%%%%%%%% Fig 2 CIFAR-100
    {\includegraphics[width=0.325\textwidth]{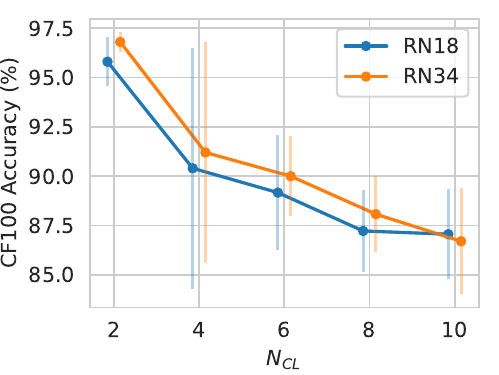}}
    \subfigure[Classification on Food-101 with two ResNets.]     %%%%%% Food-101
    {\includegraphics[width=0.325\textwidth]{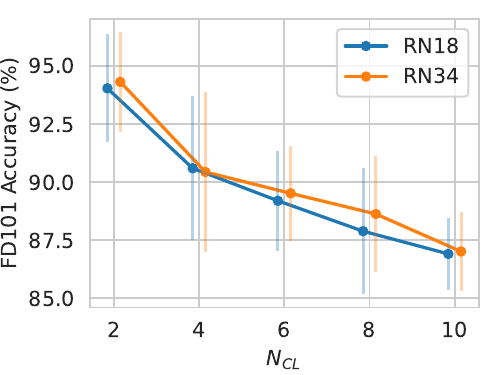}}
    \subfigure[Classification on CalTech-256 by 3 models.]
    {\includegraphics[width=0.325\textwidth]{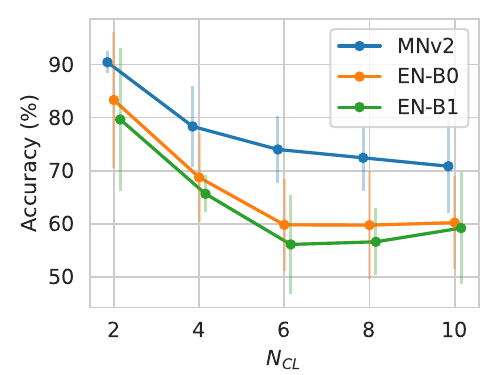}}    %%%%% 3 models on CalTech
%    \subfigure[Classification on CalTech-256 with two ResNets.]    %%% Fig 4
%    {\includegraphics[width=0.325\textwidth]{figs/resnet_vgg_v2_caltech256.pdf}}
    \\
    \subfigure[Classification on CIFAR-10 with 3 models.]        %%%%%%%%%% Fig 1 CIFAR-10
    {\includegraphics[width=0.325\textwidth]    {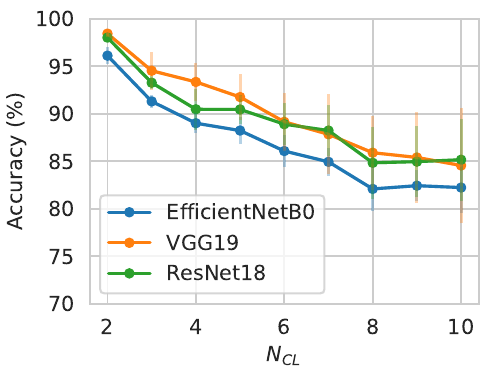}}
    \subfigure[Classification by transformers on CalTech-256.]        %%%%%%%%%% Fig 5 CalTech-256
    {\includegraphics[width=0.325\textwidth]{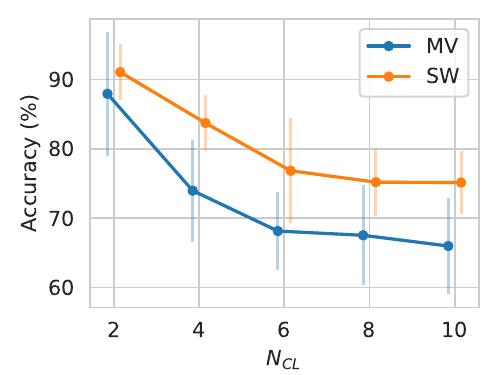}}
  \subfigure[Object detection with YOLOv5-nano on COCO.]        %%%%%%%%%% Fig 6 object detect
     {\includegraphics[width=0.325\textwidth]{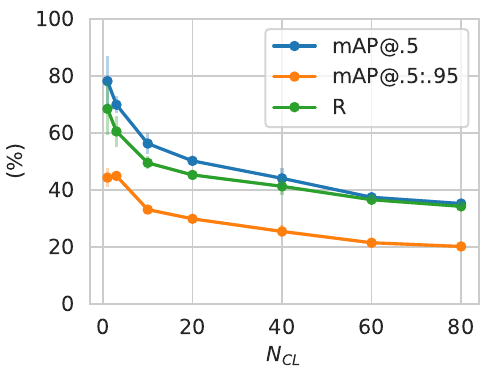}}
  \caption{{Accuracy results versus number of classes. Graphs a-d are for classification by 5 CNNs on 4 datasets. Graph e is for 2 Transformers on 1 dataset. Graph 3 is object detection by YOLOv5-nano on the COCO dataset. Each dot denotes the average accuracy of 5 subsets. The error bars show standard deviations of accuracy over 5 subsets. The CNN models are: ResNet (RN), MobileNet V2 (MNv2), and EfficientNet (EF). The transformers are: MobileViT (MV) and Swin Transformer (SW). The datasets are: CalTech-256 (CT256), CIFAR-10 (CF10), CIFAR-100 (CF100), and Food101 (FD101).}}
  \label{fig:6graphsFewClass}
  \end{minipage}
\end{figure*}

The experiments with the ResNet models and ImageNet dataset probed deeply with respect to number of models and number of subsets of classes. To obtain broader evidence of few-class properties, we performed  experiments on other models and datasets and show the results in Fig. \ref{fig:6graphsFewClass}. The results in general confirm the observed few-class properties extend over the various combinations of models and datasets tested. 

% Original >>
\clearpage
\clearpage
% Original <<

For the classification experiments of Figure \ref{fig:6graphsFewClass} a-c and e, the number of classes $N_{CL}$ is varied from $2$ to $10$ with step size $2$. For each $N_{CL}$, 5 subsets are randomly selected (seed number 0-4) from the full dataset. A model is trained and converged in each subset, then the validation Top-1 accuracy is reported. CNN models include: ResNet18 and ResNet34 \citep{he2016deep}, MobileNet V2 MobileNet V2 \citep{sandler2018mobilenetv2}, and EfficientNet-B0 and -B1 \citep{tan2019efficientnet}. By default, we use SGD optimizer with learning rate $0.1$, momentum $0.9$ and weight decay $0.0001$ to train our models. In graph f, we show the results of object detection using YOLOv5-nano on the COCO dataset for $N_{CL}$ from 80 down to 2 classes.

For graph d, image classification is performed by three models, EfficientNet-B0 \citep{tan2019efficientnet}, VGG-19 \citep{vggNet2014}, and ResNet18 \citep{he2016deep} on the CIFAR-10 dataset \citep{krizhevsky2009learning}. With so few classes (10) in this dataset -- ranging only within our self-defined few-class regime -- we wanted to see if this made any difference to the accuracy relationship. We extracted subsets of classes for $N_{CL}$ ranging from 2 to 9. We trained 3 classifiers for each subset. Results show that accuracy improvement is substantial as $N_{CL}$ reduces from 8 to 2; between 8 and 10 the accuracy curve is relatively flatter.

Graph e shows the results of classification using two transformers, MobileViT \citep{mehta2021mobilevit} and Swin Transformer \citep{liu2021swin}. Transformers are included for completeness, though they were initially designed for scaling up, i.e., larger model sizes. Since large models are not the focus of this paper, we include the lightweight ViT MobileViT.

For graph f, we decided to complement the classification experiments with an object detection experiment. We use the YOLOv5-nano ~\citep{glenn_jocher_2021_5563715} backbone upon randomly-chosen, increasing-size class subsets of the COCO dataset ~\citep{lin2014microsoft}. {Ten subsets with $N_{CL}$ of $\{$1, 2, 3, 4, 5, 10, 20, 40, 60, 80$\}$ were prepared. For each group, we trained a separate YOLOv5-nano model from scratch. The initial learning rate was set to $0.01$ with weight decay $0.0005$ at image size $640$ using SGD optimizer.} The results are similar to those of the classifiers tested, that is, the accuracy curve is steep in the few-class regime and is flatter for $N_{CL}$ from 10 to 80.

The importance of the findings in this section should be emphasized. As the number of classes is reduced, the accuracy gain for few-classes is much steeper than for many classes. This means that a practitioner with few classes can use this observation to achieve higher accuracy or a smaller model to achieve a more efficient system. We give examples of how one can take practical advantage of this in Section \ref{sec:InPractice}.

%--------------------------------------------------------------

\subsection{Intra- and Inter-Class Similarity}
\label{sec:Exp_intersim}

We start with a simple, didactic experiment to help reinforce our intuition of closeness of statistical clusters and to associate this quantitatively with the inter-similarity measure of equation \ref{eqn:S2Avg}. From the CIFAR-10 dataset, we chose 6 class groupings, three whose class features are visually similar and three whose class features are visually dissimilar. Our purpose is to examine the relationship between our visual choices, the inter-similarity measurement, and accuracy scores for three classifiers, EfficientNet-B0, VGG-19, and MobileNet-V2.

Table \ref{tab:group_sim} shows results for groupings of 2 and 4 classes from the CIFAR-10 dataset. We assign subjective assessments of visual similarity for each group; for instance, \{Deer, Horse\} are similar because they are both 4-legged animals, and \{airplane, frog\} are dissimilar because one is a large machine and another a small amphibian. One can see that the quantitative inter-class similarity $S_E$ values correspond to visual similarity ``yes-no'' assignments, high values for similar groupings and low for dissimilar. One can also see that high $S_E$ values correspond to low accuracy results from each of the models. This is as expected, the more similar are the classes, the more difficult is the classification.

\begin{table*}[ht]
\footnotesize
  \centering
    \caption{{Relationship between visual class groupings, similarity, inter-class similarity $S_E$, and accuracy for three classifiers, EfficientNet-B0 (EB0), VGG-19 (V19), and MobileNet-V2 (MV2).}}
  \label{tab:group_sim}
  \begin{tabular}{@{}c c c c c c c}
   \toprule
    $N_{CL}$ & Classes & Similar & $S_E$ & EB0 & V19 & MV2 \\
    % \specialrule{.17em}{.05em}{.05em} '
   \specialrule{.10em}{.05em}{.05em}
% \hline \hline
    4 & cat, deer, dog, horse & yes  & 0.57 & 0.84 & 0.86 & 0.76\\
    \midrule
    4 & airplane, cat, auto, ship & no & 0.12 &  0.91 & 0.94 & 0.93\\
    % \hline \hline
   \specialrule{.10em}{.05em}{.05em} 
    2 & deer, horse & yes & 0.61 & 0.92 & 0.94 & 0.89 \\
    2 & auto, truck &  yes  & 0.56 & 0.91 & 0.95 & 0.93\\
    \midrule
    2 & airplane, frog & no  & 0.11 &  0.98 & 0.98 & 0.96\\
    2 & deer, ship & no  & 0.08 & 0.98 & 0.98 & 0.96\\
   \bottomrule
  \end{tabular}
\end{table*}

\begin{figure}[t]
  \centering
    {\includegraphics[width=0.6\textwidth]{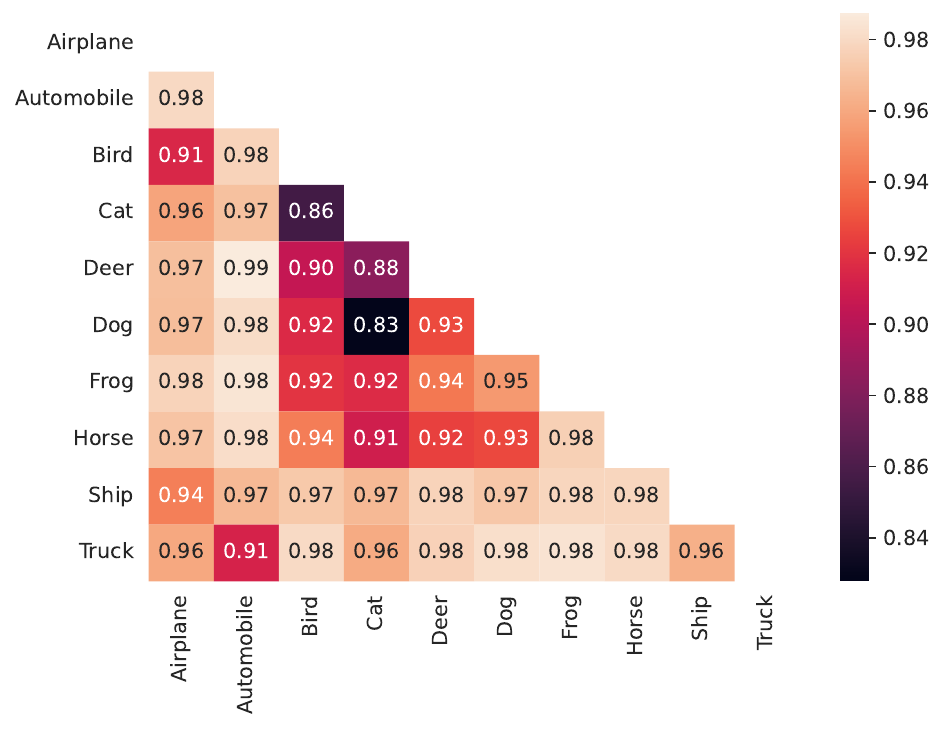}}
  \caption{Intra-class similarity values between CIFAR-10 class pairs are on the diagonal of the matrix, and inter-class values are in off-diagonal boxes. Light colored boxes indicate higher values and dark indicate lower values.}
  \label{fig:sim_matrix-sim}
\end{figure}

Just as high inter-class similarity makes classification \textit{more} difficult, high intra-class similarity makes classification \textit{less} difficult. We expand from the previous experiment to examine both intra- and inter-class similarity values for all class pairs in CIFAR-10, and give accuracy results with EfficientNet-B0 to associate with these measures. In Fig. \ref{fig:sim_matrix-sim}, intra- and inter-class similarities are given for all pairings. One of the highest intra-class similarity values is for horse, one reason may be because many of these images include a standing horse on a grassy field (i.e., low varialibity of foreground and background. Conversely, the lowest intra-class similarity value is for cat, likely because of the many poses and backgrounds. 

\begin{figure}[t]
  \centering
    {\includegraphics[width=0.6\textwidth]{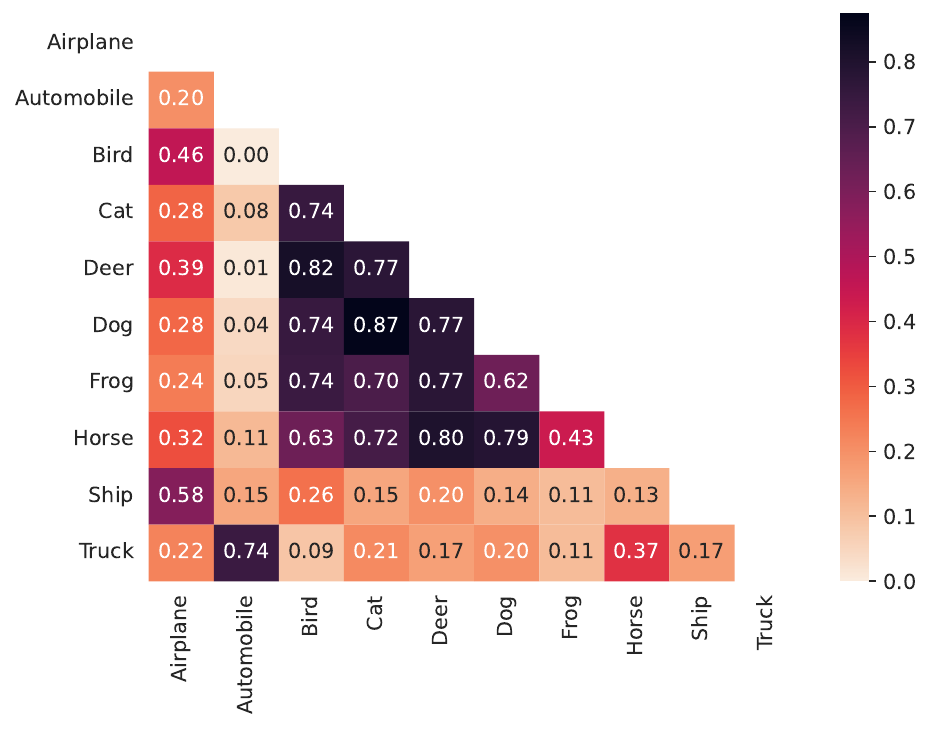}}
  \caption{EfficientNet-B0 accuracy values for CIFAR-10 class pairs. Light colored boxes indicate lower accuracies, and dark indicate higher accuracies.}
  \label{fig:sim_matrix-acc}
\end{figure}

\begin{figure}[t]
  \centering
    {\includegraphics[width=0.6\textwidth]{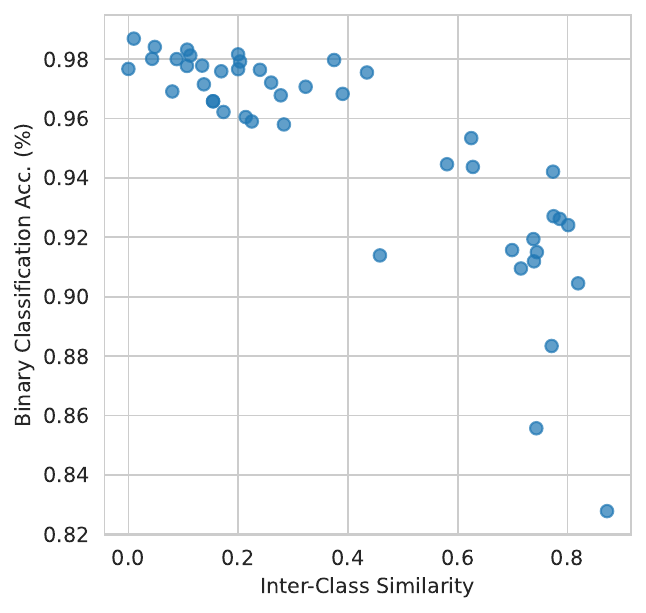}}
  \caption{As inter-class similarity increases, accuracy decreases.}
  \label{fig:sim-plot-acc}
\end{figure}

Fig. \ref{fig:sim_matrix-acc} shows EfficientNet-B0 accuracy results corresponding with each pair in Fig. \ref{fig:sim_matrix-sim}. There is a strong inverse correlation, 0.77, between similarity and accuracy values of these two graphs. Fig. \ref{fig:sim-plot-acc} shows this. Of note in this plot, the lowest similarity data point in the top left of the plot (high accuracy) is for the (automobile, deer) pair, and the highest similarity data point in the bottom right (low accuracy) is for the (cat, dog) pair.

% One by one >>

% One by one <<

% One >>
% \clearpage
% \begin{figure}[t]
%   \centering
%   % \begin{subfigure}[]{}
%     \includegraphics[width=0.32\textwidth]{figs/4.3_bi_class_EfficientNetB0_bl.pdf}
%     % \caption{Intra-class and inter-class similarity values.}
%     \label{fig:sim_matrix-sim}
%   % \end{subfigure}
%   % \hfill
%   % \begin{subfigure}[]{}
%     \includegraphics[width=0.32\textwidth]{figs/4.3_confusion_matrix_4_plot_MinMax_normed_rv_bl.pdf}
%     % \caption{Accuracy values for class pairs.}
%     \label{fig:sim_matrix-acc}
%   % \end{subfigure}
%   % \hfill
%   % \begin{subfigure}[]{}
%     \includegraphics[width=0.32\textwidth]{figs/4.3_confusion_matrix_4_plot_MinMax_normed_bi_class_EfficientNetB0_rel_v2.pdf}
%     % \caption{Accuracy decreases with increasing similarity.}
%     \label{fig:sim-plot-acc}
%   % \end{subfigure}

%   \caption{(a) Similarity values between CIFAR-10 class pairs, (b) EfficientNet-B0 accuracy matrix, and (c) Accuracy-similarity relationship. Light colors in (a) indicate higher similarity; in (b) they indicate lower accuracy.}
%   \label{fig:combined-sim-acc}
% \end{figure}
% \clearpage
% One <<

\begin{figure}[t]
\centering
    \includegraphics[width=0.65\linewidth]{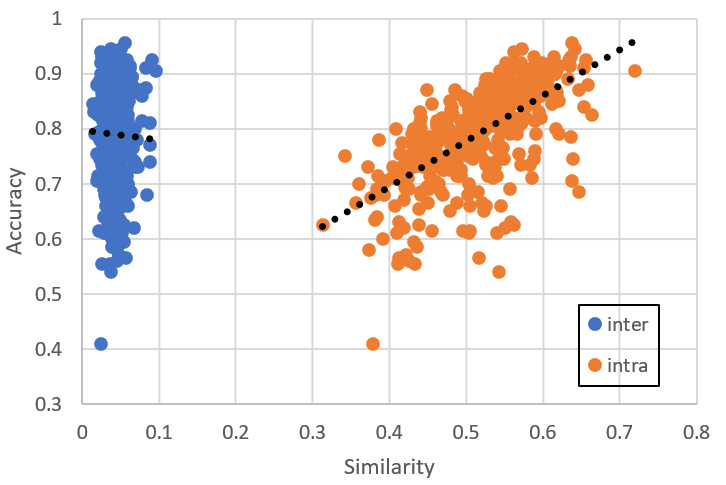}
   \caption{Clusters for accuracy versus inter-class similarity (blue) and intra-class (orange) for ResNet-50 on ImageNet-200 with 4 classes. This shows little correlation (-0.027) for inter-class, but high (0.643) for intra-class, where the dotted lines are the regression fits associated with these correlation coefficients.}
   \label{fig:intraInterVsAcc}
\end{figure}

To gain further insight into the characteristics of intra- and inter-class similarity, we augmented the few-class dataset testing of CIFAR-10, with the more extensive testing of Section \ref{sec:difficMeasure}. For each of the three models, two datasets, and 5 levels of numbers of classes, we determined the correlation between accuracy and both intra- and inter-class similarity measures for each class grouping of 200 class combinations. We show a representative plot the intra- and inter-class clusters of similarity versus accuracy in Fig. \ref{fig:intraInterVsAcc}. This plot is for the ResNet-50 model on the ImageNet-200 dataset and for $N_{CL}  = 4$; the other 29 model-dataset-$N_{CL}$ combinations (3 models $\times$ 2 datasets $\times$ 5 numbers of classes - 1) have similar characteristics. Correlation values for combinations with the ResNet-50 model are shown in Table \ref{tab:group_sim}; correlations for MobileNet-V2 and ViT are not shown, but also have similar characteristics.

The plot of Fig. \ref{fig:intraInterVsAcc} illustrates some important characteristics on intra- and inter-class similarity. The most evident is that inter-class similarity values are much smaller than intra-class values because different classes have very different features than those in the same class. However, most interesting is that the correlation of intra-class similarity to accuracy is much greater than for inter-class, as illustrated by the dashed lines in Fig. \ref{fig:intraInterVsAcc} and by the correlation values in Table \ref{tab:accVsSim}. This is also reinforced in Fig. \ref{fig:RVsNc6plots} where the highest correlation of difficulty score to accuracy is for high weighting of the intra-class similarity measure and much lower for inter-class similarity. This means that close clustering within classes, as measured by intra-class similarity is a much better predictor of accuracy than distance between clusters as measured by inter-class similarity -- at least for the datasets tested. We examined our random class combinations to understand this better. The three top inter-class similarities for two classes are for (lion, tiger), (bowl, plate), and (shark, shrew). But there were very few combinations such as these in which the two classes were so obviously similar. And as the number of classes increased above two, both intra- and inter-class similarity tend toward their respective averages as there is less likelihood of close similarities for more classes. 

% Pre
% \begin{table}[ht]
%   \centering
%     \caption{{Correlation of accuracy versus similarity for intra- and inter-class similarities for the ResNet-50 test.}}
%   \label{tab:accVsSim}
%   \begin{tabular}{@{}c | c  c c c      c c}
%    &&\multicolumn{2}{c}{\textbf{ImageNet}} && \multicolumn{2}{c}{\textbf{CIFAR}}\\
% %    $N_{CL}$ & ImageNet &  & CIFAR & \\
%     $N_{CL}$ & & intra & inter & & intra & inter \\
%    \hline
% 2 & & 0.585	& 0.060	& & 0.807	& -0.097 \\
% 3 & & 0.635	&0.041	& & 0.826	&-0.210\\
% 4 & & 0.643	&-0.027	& & 0.819	&-0.238\\
% 6 & & 0.422	&0.031	& & 0.801	&-0.300\\
% 8 & & 0.557	&-0.055	& & 0.833	&-0.365\\
%   \end{tabular}
% \end{table}

\begin{table}[ht]
\footnotesize
  \centering
    \caption{{Correlation of accuracy versus similarity for intra- and inter-class similarities for the ResNet-50 test.}}
  \label{tab:accVsSim}
  \begin{tabular}{@{}c  c  c c c      c c}
  \toprule
   &&\multicolumn{2}{c}{\textbf{ImageNet}} && \multicolumn{2}{c}{\textbf{CIFAR}}\\
   
%    $N_{CL}$ & ImageNet &  & CIFAR & \\
    $N_{CL}$ & & intra & inter & & intra & inter \\
   \midrule
2 & & 0.585	& 0.060	& & 0.807	& -0.097 \\
3 & & 0.635	&0.041	& & 0.826	&-0.210\\
4 & & 0.643	&-0.027	& & 0.819	&-0.238\\
6 & & 0.422	&0.031	& & 0.801	&-0.300\\
8 & & 0.557	&-0.055	& & 0.833	&-0.365\\
    \bottomrule
  \end{tabular}
\end{table}

% \vspace{-10pt}

\newpage
%--------------------------------------------------------------
\subsection{Class Clustering}

The intra- or inter-class similarity alone neglects cluster characteristic information that can better represent the a dataset difficulty. Specifically, this dataset property should capture the \textbf{(1) tightness of a class cluster} and \textbf{(2) distance to other classes}. To that end, we adopt the Silhouette Score (SS) \citep{rousseeuw1987silhouettes, shahapure2020cluster}: $SS(i) = \frac{b(i) - a(i)}{max({a(i), b(i)})}$,
where $SS(i)$ is the Silhouette Score of the data point $i$, $a(i)$ denotes the average dissimilarity between $i$ and other instances in the same class, and $b(i)$ represents the average dissimilarity between $i$ and other data points in the closest different class.
% Intuitively, this metric summarizes the quality of clusters jointly by the degree of instances of the same class and distinct clusters, normalized by the longest distance of $a(i)$ and $b(i)$. By this definition, we can see that $-1 \leq SS(i) \leq 1$ where $-1$ indicates a dataset is poorly clustered (data points with different classes are broadly scattered) while $1$ represents a closely-clustered dataset.
% (also called Silhouette Coefficient in the literature)

% Euclidean Distance is commonly used to measure two data points' differences; in contrast, we incorporate the inverse of similarity (dissimilarity) as data points' differences into the existing Silhouette Score.

% Observe that the above Intra-Class Similarity $S_{\alpha}^{(C)}$ already represents the tightness of the class $(C)$, therefore $a(i)$ can be replaced with the inverse of Intra-Class Similarity $a(i) = -S_{\alpha}(i)$. For the second term $b(i)$, we adopt the previously defined Inter-Class Similarity $S_{\beta}^{(C_{1}, C_{2})}$ and introduce a new similarity score as \textbf{Nearest Inter-Class Similarity} $S^{\prime}_{\beta}^{(C)}$, which is a scalar describing the similarity among instances between class $C$ and the closest class of each instance in $C$. 

% \newpage
We denote the dataset-level Nearest Inter-Class Similarity ${S^{\prime}}_{\beta}^{(D)}$ as:
% \bo{Change hat into prime. Follow the notations in the sillouette paper.}
% \begin{equation}
%     {S^{\prime}}_{\beta}^{(C)} = \frac{1}{\vert P^{(C, \hat{C})} \vert}\sum_{i \in C, j \in \hat{C}}\cos(\textbf{Z}_{i}, \textbf{Z}_{j}),
% \end{equation}
% where $\hat{C}$ is the set of the nearest class to $C$ ($\hat{C} \neq C$).
% Consequently, the dataset-level Nearest Inter-Class Similarity ${S^{\prime}}_{\beta}^{(D)}$ is expressed as:

% \larry{This equation looks good to me, but there is a LaTeX error.}
\begin{equation}
\begin{aligned}
    {S^{\prime}}_{\beta}^{(D)} &= \frac{1}{\vert L\vert} \sum_{l \in L} {S^{\prime}}_{\beta}^{(C_{l})} \\
    &= \frac{1}{\vert L\vert \times \vert P^{(C_{l}, \hat{C_{l}})} \vert} \sum_{l \in L} \mspace{5mu} \sum_{i \in C_{l}, j \in \hat{C_{l}}}\cos(\textbf{Z}_{i}, \textbf{Z}_{j}),
\end{aligned}
\end{equation}
where $\hat{C}$ is the nearest class to instance $i$ ($\hat{C} \neq C$).
% The second term of $SS(i)$ can be written as $b(i) = -{S^{\prime}}_{\beta}(i)$.
% Replacing $a(i)$ and $b(i)$ from equation \ref{equ:SS} with these similarity terms, 

% \larry{Again, check the LaTeX errors, though text looks fine.}
Taking all into consideration, we propose a novel \textbf{Similarity-Based Silhouette Score $SimSS$} for dataset $D$:
\begin{equation}
    \begin{split}
    SimSS^{(D)} &= \frac{1}{\vert L \vert \times \vert C_{l} \vert}\sum_{i \in C_{l}}\frac{S_{\alpha}(i) - {S^{\prime}}_{\beta}(i)}{max(S_{\alpha}(i), {S^{\prime}}_{\beta}(i))}.
    \end{split}
    \label{equ:simss_D_main}
\end{equation}

\begin{figure}[h]
  % \begin{minipage}{1\linewidth}
  \centering
    % \subfigure[]{\includegraphics[width=0.85\textwidth]{figs/sim_line_clip.pdf}}
    % \subfigure[]{\includegraphics[width=0.85\textwidth]{figs/sim_line_dinov2.pdf}}
    % \subfigure[]{\includegraphics[width=0.85\textwidth]{figs/sim_line_clip_v1.1.pdf}}
    % \subfigure[]{\includegraphics[width=0.85\textwidth]{figs/sim_line_dinov2_v1.1.pdf}}
    % \subfigure[]{\includegraphics[width=0.85\textwidth]{figs/sim_line_clip_v1.2.pdf}}
    % \subfigure[]{\includegraphics[width=0.85\textwidth]{figs/sim_line_dinov2_v1.2.pdf}}
    % \subfigure[SimSS using CLIP as similarity base function vs $N_{CL}$ curve.]{\includegraphics[width=0.85\textwidth]{figs/sim_line_clip_v1.3.pdf}}
    % \subfigure[SimSS using CLIP as similarity base function vs $N_{CL}$ curve.]{\includegraphics[width=0.85\textwidth]{figs/sim_line_clip_v1.4.pdf}}
    \subfigure[SimSS using CLIP as similarity base function vs $N_{CL}$ curve.]{\includegraphics[width=0.496\textwidth]{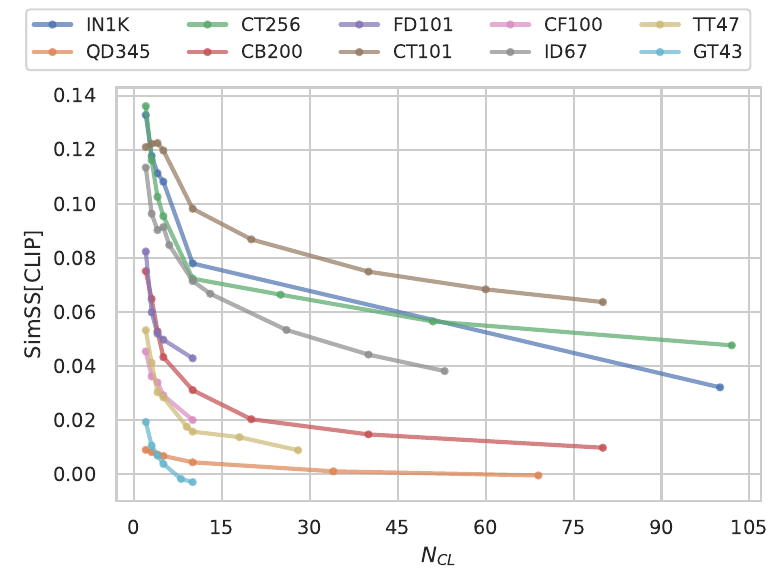}}
    % \subfigure[SimSS using DINOv2 as similarity base function vs $N_{CL}$ curve.]{\includegraphics[width=0.85\textwidth]{figs/sim_line_dinov2_v1.3.pdf}}
    % \subfigure[SimSS using DINOv2 as similarity base function vs $N_{CL}$ curve.]{\includegraphics[width=0.85\textwidth]{figs/sim_line_dinov2_v1.4.pdf}}
    \subfigure[SimSS using DINOv2 as similarity base function vs $N_{CL}$ curve.]{\includegraphics[width=0.496\textwidth]{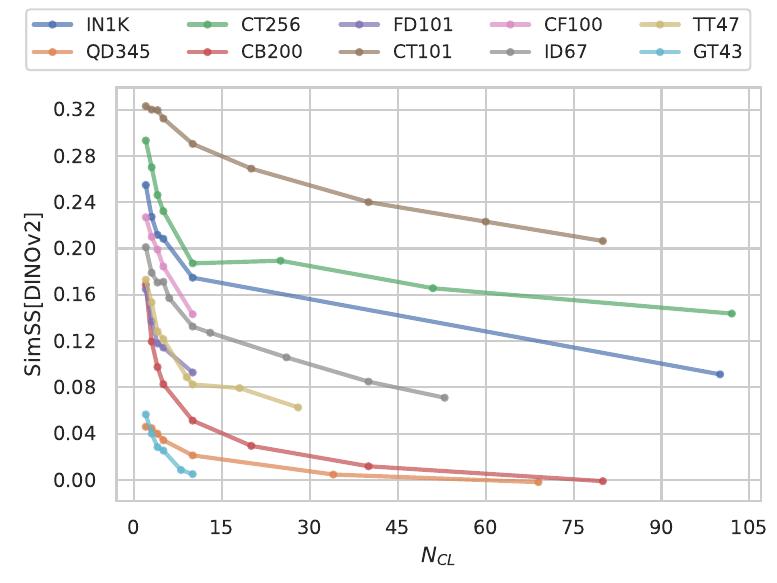}}
  \caption{Relation of SimSS[CLIP,DINOv2] and $N_{CL}$.}
  % \bo{TODO: thinner without dots.}
  \label{fig:bump_10m10ds_v3p2p1}
  % \end{minipage}
\end{figure}

Using our proposed SimSS, the relationships of similarity scores and number of classes $N_{CL}$ in ten datasets \footnote{Datasets include CalTech101 (CT101) \citep{li_andreeto_ranzato_perona_2022}~, CalTech256 (CT256) \citep{griffin_holub_perona_2022}~, CIFAR100 (CF100) \citep{krizhevsky2009learning}~, CUB200 (CB200) \citep{wah_branson_welinder_perona_belongie_2011}~, Food101 (FD101) \citep{bossard14}~, GTSRB43 (GT43) \citep{Stallkamp2012}~, ImageNet1K (IN1K) \citep{deng2009imagenet}~, Indoor67 (ID67) \citep{quattoni2009recognizing}~, Quickdraw345 (QD345) \citep{ha2017neural}~ and Textures47 (TT47) \citep{cimpoi14describing}~.} are shown in Fig. \ref{fig:bump_10m10ds_v3p2p1} (a) and (b) with CLIP and DINOv2 as similarity base functions.

In Fig. \ref{fig:bump_10m10ds_v3p2p1}, we make a key observation that all ten datasets share the general trend of inverse relationship between similarity and the number of classes. In particular, image similarity increases as the number of classes $N_{CL}$ decreases. This reveals that similarity, as a proxy of dataset difficulty, plays a more important role in the \fcr~ than for datasets with more classes. Therefore, blindly downscaling a model without considering dataset difficulty may yield a sub-optimal model selection decision. We therefore propose to take image similarity into consideration in existing scaling laws \citep{kaplan2020scaling, rae2021scaling, zhai2022scaling}.

% \clearpage
% (a): (lab) brcao@dawn:~/Repos/plots/Y2024M05D20Mon$ python3 sim_line_clip.py 
% (b): (lab) brcao@dawn:~/Repos/plots/Y2024M05D20Mon$ python3 sim_line_dinov2.py
% (a): (lab) brcao@dawn:~/Repos/plots/Y2024M05D20Mon$ python3 sim_line_clip_v1.1.py 
% (b): (lab) brcao@dawn:~/Repos/plots/Y2024M05D20Mon$ python3 sim_line_dinov2_v1.1.py
% (a): (lab) brcao@dawn:~/Repos/plots/Y2024M05D20Mon$ python3 sim_line_clip_v1.2.py 
% (b): (lab) brcao@dawn:~/Repos/plots/Y2024M05D20Mon$ python3 sim_line_dinov2_v1.2.py
% (a): (lab) brcao@dawn:~/Repos/plots/Y2024M05D20Mon$ python3 sim_line_clip_v1.3.py 
% (b): (lab) brcao@dawn:~/Repos/plots/Y2024M05D20Mon$ python3 sim_line_dinov2_v1.3.py
% (a): (lab) brcao@dawn:~/Repos/plots/Y2024M05D20Mon$ python3 sim_line_clip_v1.4.py 
% (b): (lab) brcao@dawn:~/Repos/plots/Y2024M05D20Mon$ python3 sim_line_dinov2_v1.4.py
% (a): (lab) brcao@dawn:~/Repos/plots/Y2024M05D20Mon$ python3 sim_line_clip_v1.5.py 
% (b): (lab) brcao@dawn:~/Repos/plots/Y2024M05D20Mon$ python3 sim_line_dinov2_v1.5.py
% \clearpage

\noindent \textbf{Comparing Accuracy and Similarity.}
% \vspace{-13pt}
\begin{figure}[h]
  \begin{minipage}{1\linewidth}
  \centering
    \subfigure[DCN-Full]{\includegraphics[width=0.23\textwidth]{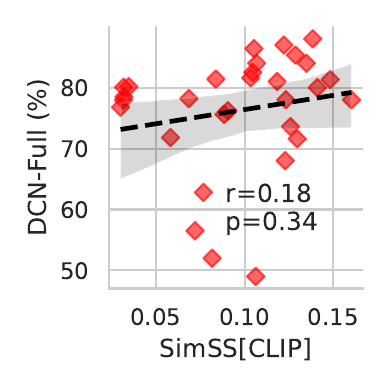}}
    % \hspace{7pt}
    \subfigure[DCN-Full]{\includegraphics[width=0.23\textwidth]{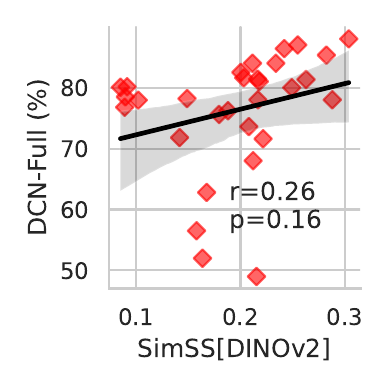}}
    % \hspace{7pt}
    \subfigure[DCN-Sub]{\includegraphics[width=0.23\textwidth]{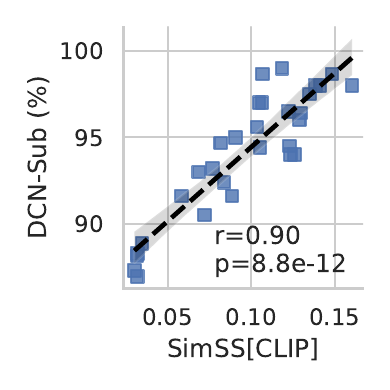}}
    % \hspace{7pt}
    \subfigure[DCN-Sub]{\includegraphics[width=0.23\textwidth]{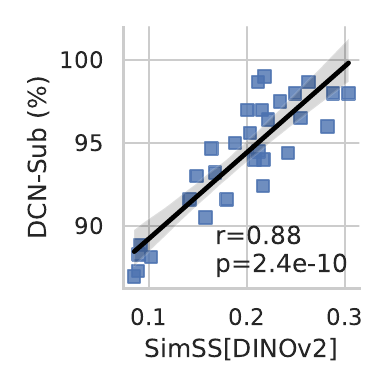}}
  % \hspace{-6pt}
  \captionsetup{font=small}
  % \vspace{-5pt}
  \caption{Pearson correlation coefficient ($r$) between DCN and SimSS when $N_{CL} \in \{2, 3, 4, 5, 10, 100\}$. DCN-Sub (blue squares) is more highly correlated than DCN-Full (red diamonds) with SimSS using both similarity base  functions of CLIP (dashed line) and DINOv2 (solid line) with $r \geq 0.88$.}
  \label{fig:simcorr}
  \end{minipage}
\end{figure}
% (a): plots/Y2024M05D20Mon/fca_full_sim_resnet_dcn_clip_dinov2.py
% (b): plots/Y2024M05D20Mon/fca_full_sim_resnet_dcn_clip_dinov2.py
% (c): plots/Y2024M05D20Mon/fca_sub_sim_resnet_dcn_clip_dinov2.py
% (d): plots/Y2024M05D20Mon/fca_sub_sim_resnet_dcn_clip_dinov2.py
% https://blogs.sas.com/content/iml/2023/04/05/interpret-spearman-kendall-corr.html
% https://www.ncbi.nlm.nih.gov/pmc/articles/PMC3576830/
% https://www.ncl.ac.uk/webtemplate/ask-assets/external/maths-resources/statistics/regression-and-correlation/strength-of-correlation.html
% https://www.statstutor.ac.uk/resources/uploaded/spearmans.pdf
% https://sphweb.bumc.bu.edu/otlt/MPH-Modules/PH717-QuantCore/PH717-Module9-Correlation-Regression/PH717-Module9-Correlation-Regression4.html
The effectiveness of SimSS is evaluated by Pearson correlation coefficient (PCC) ($r$) between model accuracy and SimSS. The reference dataset classification difficulty number (DCN) \citep{scheidegger2021efficient} is adopted to represent the empirical top accuracy in a dataset. We show that in Figure \ref{fig:simcorr} (a) (b), SimSS is poorly correlated with DCN-Full ($r=0.18$ and $r=0.26$ for CLIP and DINOv2). This could be due to the large variance accuracy and extraneous feature representations from full datasets to a much smaller set of target sub-classes. In contrast, with $r=0.90$ and $r=0.88$ using CLIP (dashed) and DINOv2 (solid), SimSS is highly correlated with DCN-Sub (blue squares). We attribute the high correlation to sub-models' focus on the minimal features needed in the target sub-classes. 
The high correlation score \citep{pcc, schober2018correlation} demonstrates that SimSS can be regarded as a reliable metric for estimating few-class dataset difficulty. Such a metric can help predict a model's empirical upper-bound accuracy. We highlight that this score is computed only once and then used for all times for the same dataset.

%--------------------------------------------------------------

\subsection{Other Features: Color and Scale-Resolution}
\label{sec:Exp_color}
Because reduction from color to grayscale only affects the number of multiplies in the first layer, the efficiency gain depends upon the number of layers of the model. For a large model such as VGG-19, percentage efficiency gain will be much smaller than for a small model such as EfficientNet-B0, as shown in Table \ref{tab:colorComparison}. Furthermore, even for small networks, the effect of reducing from color to grayscale processing is small relative to effects of other attributes. Because of these small gains, we do not pursue the investigation of color-to-grayscale efficiencies further. For a previous investigation of color versus gray neural network computation, refer to \citep{grayscaleBetter2016}.

% Pre
% \begin{table}[ht]
%   \centering
%     \caption{ Difference between color and grayscale computation [kFLOPS] for the large VGG-19 classifier and much smaller EfficientNet-B0 classifier. Ratio is grayscale-to-color computation for all layers. 
%   }
%   \label{tab:colorComparison}
%   % \begin{tabular}{c|c|c|c|c|c}
% %  \begin{tabular}{p{35} | p{27} p{27} | p{27} p{27} | p{20}}
%   \begin{tabular}{c |c c | c c | c}
  
%   Model & \multicolumn{2}{c}{Layer-1} & \multicolumn{2}{c}{All Layers} & Ratio \\
%  % \cmidrule{2-6}
%      & color & gray & color & gray & [\%]\\
%    \midrule
%     VGG-19 & 1835.01 & 655.36 & 399474 & 398295 & 99.7 \\
%     EN-B0 & 884.7 & 294.9 & 31431 & 30841 & 98.1 \\
%   \end{tabular}
% \end{table}

\begin{table}[ht]
\footnotesize
  \centering
    \caption{ Difference between color and grayscale computa- tion [kFLOPS] for the large VGG-19 classifier and much smaller EfficientNet-B0 classifier. Ratio is grayscale-to- color computation for all layers. 
  }
  \label{tab:colorComparison}
  % \begin{tabular}{c|c|c|c|c|c}
%  \begin{tabular}{p{35} | p{27} p{27} | p{27} p{27} | p{20}}
  \begin{tabular}{c |p{0.8cm} p{0.8cm}| p{0.8cm} p{0.8cm} | c}
  \toprule
  Model & \multicolumn{2}{c}{Layer-1} & \multicolumn{2}{c}{All Layers} & Ratio \\
 % \cmidrule{2-6}
     & color & gray & color & gray & [\%]\\
   \midrule
    VGG-19 & 1835.01 & 655.36 & 399474 & 398295 & 99.7 \\
    EN-B0 & 884.7 & 294.9 & 31431 & 30841 & 98.1 \\
    \bottomrule
  \end{tabular}
\end{table}

The fifth property of an image dataset is scale-resolution. We call this a single property because the scale and resolution are inter-dependent. When the size of an image is reduced through subsampling, so is pixel resolution. For efficient model selection, a practitioner would strive to reduce the image size as much as possible to reduce the model size and number of multiplications. In many cases, such as when an image is captured at a resolution that exceeds that of important features, scale reduction can yield great efficiency benefits. However, scale reduction can lead to accuracy loss when smaller-scale features such as texture, object boundaries, distance between objects, etc., contributes to class separation. There is substantial emphasis on the model side to be scale efficient. For instance, model families such as EfficientNet contain increasingly large models with increasing scale steps. (For EfficientNet, the levels and input sizes (one dimension) are: B0 (224), B1 (240), B2 (260), B3 (300), B4 (380), B5 (456), B6 (528), B7 (600).) On the data-side, we know of no previous work that examines scale-resolution versus accuracy and measures difficulty as we have done for the other data-side properties. While object scale is straightforward to quantify, feature resolution (not pixel resolution) is more difficult. We leave the analysis of the fifth data-side property, scale-resolution, to future work.

%---------------------------------------------------------------------

\section{Few-Class Models in Practice}
\label{sec:InPractice}

How can a practitioner take advantage of their knowledge of properties of their few-class application to more easily design efficient applications? We describe five aspects, i) sub-models that offer extra efficiency for few-class applications, ii) fast selection of efficient models using the difficulty metric, iii) dataset difficulty comparison, iv) design-phase aid for specifying an application, and v) embedded systems examples.

\subsection{Sub-Models}
Practitioners with few-class applications and need for efficiency will often choose the smallest model from a model family, for example YOLO-nano from the YOLO family (nano to x-large) or EfficientNet-B0 from the EfficientNet family (B0 to B7). We have found empirically that, using the same compound scaling factor used to increase width, depth, and resolution of a family for larger models, we can also scale \textit{down} to smaller models -- sometimes with little accuracy loss.

For the YOLOv5 object detector, we scale layers and channels down in model size with the depth and width factors already used for scaling the family up in size from nano to x-large \citep{glenn_jocher_2021_5563715}. Starting with depth and width multiples of 0.33 and 0.25 for YOLOv5-nano, we reduce these in step sizes of 0.04 for depth and 0.03 for width. In this way, we design a monotonically decreasing sequence of sub-YOLO models that we denote as SY1 to SY8. 

We prepared 89 random class subsets from the COCO minitrain dataset ~\citep{HoughNet}. These include 80 subsets with $N_{CL}=1$, each containing a single class from 80 classes; 8 subsets with $N_{CL}=10$; and the final dataset is the original COCO minitrain with $N_{CL}=80$. We trained each model separately for each of these groupings.

\begin{figure}[t]
  \centering
    {\includegraphics[width=0.55\linewidth]{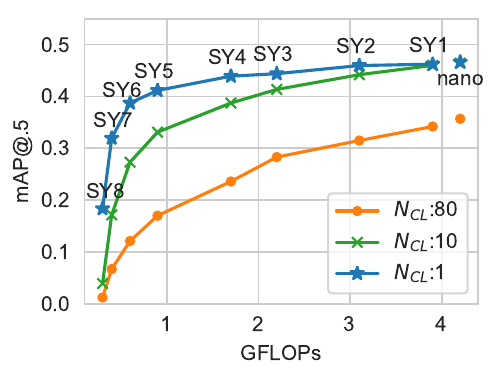}}
  \caption{Computation reduction (GFLOPs) as model size is reduced from YOLO-nano to sub-models SY1 to SY8.}
  \label{fig:subYOLO_GFLOPs}
\end{figure}

Results of sub-YOLO detection are shown in Fig. \ref{fig:subYOLO_GFLOPs}. There are three plots in which each point of $mAP@.5$ is averaged across results for all subsets for each $N_{CL}$. An overall trend can be observed that the fewer-class models achieve higher efficiency (lower GFLOPs) than many-class models. This is as expected, but perhaps not expected and certainly beneficial to few-class model application developers is the following. Whereas the accuracy for 80 classes drops steadily from the YOLOv5-nano size, accuracy for 10 classes is fairly flat down to SY2, which corresponds to a 36\% computation reduction, and for 1 class down to SY4, which corresponds to a 72\% computation reduction.

There are three messages from this experiment to practitioners with few-class applications. One is that smaller models may be available that provide greater efficiency at similar accuracy than the smallest (published) models of a family. The second message is that the smaller models do not have to be architected from scratch; instead, the compound scaling factor for the family can be used to down-scale the same model architecture by simply adjusting width, depth, and resolution parameter values in the model configuration. Thirdly, the performance advantage of sub-models is greater for few-classes datasets than larger.

\subsection{Selecting Models Using the Difficulty Metric}
\label{sec:cmpModels}

To select a neural network model for an application, a common first step is to identify models that might be appropriate to meet the requirements. In this paper, since we focus on small, efficient computer vision applications, appropriate selections would include scaled model families such as YOLO, MobileNet, and EfficientNet-B0 (either classification or object detection variants of these depending upon the goal of the application). The traditional selection process entails training different model sizes and families and comparing these. This is a tedious and time-consuming process if the practitioner does (as is one of the messages of this paper to obtain an efficient model) train each model candidate on each application dataset.

Use of the difficulty metric offers a faster way to find a well-matched model. The practitioner follows the following procedure. First, training is performed on a dataset that includes all classes of interest (and it can include other classes that are not of immediate interest). The choice of training model doesn't matter since we will be comparing groups of classes relative to their features, i.e., data-side features, rather than comparing between models, i.e., model side. This is done once, and although it can be done once by a practitioner, it is more logically done and offered as a public benchmark. For this purpose, we have created a "few-class benchmark" \citep{cao2025few}, which provides difficulty and accuracy results for many combinations of classes, number-of-classes, models, and datasets\footnote{Code is available at {https://github.com/bryanbocao/fca}}. This benchmark eliminates the need for a practitioner to perform the initial training step -- as long as the classes in the application are also in the benchmark. We leave description of the benchmark to the reference.

With use of the trained model and equation
\ref{eqn:DifficF} (with $\gamma = 0.25$), difficulty measures can be calculated for all class groupings of interest. By use of the difficulty metric alone, this gives the practitioner the numbers upon which to perform prediction of \textit{relative} performance between groupings. After choosing classes, number of classes, and model, that combination of class grouping and model is trained and tested once to predict absolute (not relative) accuracy performance.

Following is a simple application example. 
Consider a computer vision application to monitor safety items worn on a construction site. A dataset of construction safety wear is created and trained that includes: helmet, reflective vest, protective footwear, protective eyewear, gloves, lumbar support, etc. A practitioner can compare the difficulties of any combination of these objects  -- \textit{without the need for training and testing each grouping}. After a grouping and model are chosen, that model is trained and tested once to predict performance, and used for the application.

An experimental comparison of use of the difficulty metric for comparing models is shown in Table ~\ref{tab:sim_eff}. For the CIFAR-10 dataset, we compare three CNNs, VGG-19, EfficientNet-B0, and MobileNet-V2, each designed for small or embedded \citep{roth2024resource} applications, on a NVIDIA RTX A6000 GPU. Both CNN and difficulty metric approaches require initial training. For the difficulty metric, there is an additional one-time task of feeding all test images by class into the difficulty metric, which takes 0.72 seconds, and then caching, which takes 0.63 seconds. This is included in the difficulty metric training time of 3.31 in the table. This time is larger than for each CNN, but there are two qualifications to make with this. One is that training is done only once (versus testing times in the next column, which are done for each instance pair). The second qualification is that, when the difficulty metric is found in a public benchmark, the practitioner incurs no training time.  Applying the difficulty metric takes 0.76 seconds to calculate a difficulty score for each instance pair. The conventional alternative is full CNN testing, which as seen in the Table, takes from 6 to 20 times longer. 

\begin{table}[ht]
\footnotesize
  \centering
    \caption{{Runtime comparison results for conventional CNN training and testing versus using the difficulty metric. s: second, pair: all pair of instances in two classes.}}
  \label{tab:sim_eff}
%\begin{tabular}{@{}l p{50} p{40}}
\begin{tabular}{p{95pt}|  p{45pt} p{45pt}}
    \toprule
    Model & $t_{train}$ & $t_{test}$ \\
          & (s/epoch) & (s/pair) \\
    \midrule
    % VGG-19 & 0.692 & 4.49 \\
    VGG-19 & 0.69 & \hspace{1pt} 4.49 \\
    EfficientNet-B0 & 3.13 & 21.82 \\
    MobileNet V2 & 2.19 & 15.05 \\
    % SM & 3.305 & 0.762 \\
    \textbf{Difficulty Metric} & 3.31 & \hspace{-1pt} \textbf{0.76} \\
    \bottomrule
  \end{tabular}
\end{table}
% \vspace{-20px}

%--------------------------------------------------%

\subsection{Dataset Difficulty}
\label{sec:datasetDiffic}
Although we used the term \textit{classification difficulty}, in Section \ref{sec:difficMeasure}, a narrower use of equation \ref{eqn:DifficF} can be termed \textit{dataset difficulty}. We use class groupings from the experiments in Section \ref{sec:difficMeasure} to illustrate how dataset difficulty relates to difficulty. In Table \ref{tab:DatasetDiffic}, the lowest (L) and highest (H) dataset difficulty values are found and shown with their corresponding accuracies and class grouping. For $N_{CL}=2$, it is especially evident that (leopard, tiger), (poppy, tulip), and (bowl, plate) each comprise similar class pairs; their difficulty measures are high and accuracies low as expected. It is less easy to recognize that (orange, tank), (bicycle, chair), and (skyscraper, tiger) are the least difficult pairs; and even less easy for $N_{CL}=4$. The point of this example is that, given a dataset, the difficulty metric can be used as a quick prediction (i.e., without repeated training and testing) of the relative performance, or equivalently the relative resources needed to perform classification upon compared datasets.

% Pre
% \begin{table}[ht]
%   \caption{Dataset difficulty scores and corresponding accuracies on CIFAR-100 with MobileNet (MN), ResNet (RN), and ViT. L and H denote the 3 lowest and highest difficulty values from 200 sub-groups for $N_{CL}=\{2,4\}$ each.}
%   \label{tab:DatasetDiffic}
%   \centering
%   \footnotesize
%   \begin{tabular}{c c | c c c | l}
%     $N_{CL}$ & Difficulty & MN & RN & ViT & Classes\\
%     \toprule
%     2 & 0.055 (L) & 0.83 & 0.92 & 0.98 & orange, tank\\
%      & 0.060 (L) & 0.85 & 0.90 & 0.97 & bicycle, chair\\
%      & 0.062 (L) & 0.78 & 0.91 & 0.95 & skyscraper, tiger\\
%      \cmidrule{2-6}
%      & 0.259 (H) & 0.69 & 0.81 & 0.93 & leopard, tiger\\
%     & 0.283 (H) & 0.67 & 0.75 & 0.88 & poppy, tulip\\
%     & 0.369 (H) & 0.63 & 0.68 & 0.83 & bowl, plate\\
%     \midrule
%     4 & 0.046 (L) & 0.81 & 0.88 & 0.96 & bicycle, bottle,\\
%     & & & & & chimp, poppy\\
%      & 0.051 (L) & 0.81 & 0.88 & 0.97 & bicycle, castle, \\
%      & & & & & chimp, mushroom\\
%      & 0.051 (L) & 0.88 & 0.90 & 0.96 & aquarium, motorcycle, \\
%      & & & & & road, streetcar\\
%      \cmidrule{2-6}
%      & 0.154 (H) & 0.53 & 0.60 & 0.81 & boy, lobster, \\
%      & & & & & man, possum\\
%     & 0.160 (H) & 0.61 & 0.70 & 0.83 & crocodile, maple tree, \\
%     & & & & & oak tree, turtle\\
%     & 0.210(H) & 0.64 & 0.73 & 0.80 & oak tree, pine tree, \\
%     & & & & & tiger, willow tree\\
%   \end{tabular}
% \end{table}

\begin{table*}[ht]
\footnotesize
  \caption{Dataset difficulty scores and corresponding accuracies on CIFAR-100 with MobileNet (MN), ResNet (RN), and ViT. L and H denote the 3 lowest and highest difficulty values from 200 sub-groups for $N_{CL}=\{2,4\}$ each.}
  \label{tab:DatasetDiffic}
  \centering
  % \footnotesize
  \begin{tabular}{c c | c c c | l}
    \toprule
    $N_{CL}$ & Difficulty & MN & RN & ViT & Classes\\
    \toprule
    2 & 0.055 (L) & 0.83 & 0.92 & 0.98 & orange, tank\\
     & 0.060 (L) & 0.85 & 0.90 & 0.97 & bicycle, chair\\
     & 0.062 (L) & 0.78 & 0.91 & 0.95 & skyscraper, tiger\\
     \cmidrule{2-6}
     & 0.259 (H) & 0.69 & 0.81 & 0.93 & leopard, tiger\\
    & 0.283 (H) & 0.67 & 0.75 & 0.88 & poppy, tulip\\
    & 0.369 (H) & 0.63 & 0.68 & 0.83 & bowl, plate\\
    \midrule
    4 & 0.046 (L) & 0.81 & 0.88 & 0.96 & bicycle, bottle, chimp, poppy\\
     & 0.051 (L) & 0.81 & 0.88 & 0.97 & bicycle, castle, chimp, mushroom\\
     & 0.051 (L) & 0.88 & 0.90 & 0.96 & aquarium, motorcycle, road, streetcar\\
     \cmidrule{2-6}
     & 0.154 (H) & 0.53 & 0.60 & 0.81 & boy, lobster, man, possum\\
    & 0.160 (H) & 0.61 & 0.70 & 0.83 & crocodile, maple tree, oak tree, turtle\\
    & 0.210 (H) & 0.64 & 0.73 & 0.80 & oak tree, pine tree, tiger, willow tree\\
    \bottomrule
  \end{tabular}
\end{table*}

\subsection{Application Specification}
\label{sec:applSpec}
When practitioners specify application requirements, they consider tradeoffs in accuracy, size, speed, power usage, and cost. We have applied the classification difficulty metric to model selection for a real industry application involving video analytics for human-robot interaction \citep{ogorman2024pathplan}. The objective is to recognize human activity from fixed hallway cameras of an assembly factory so as to reduce human-robot interaction (HRI). Three classes were identified and trained for this application, \textit{person-walk}, \textit{person-cart}, and \textit{robot}. The \textit{person} class was initially separated into two, \textit{person-walk} and \textit{person-cart} (person walking and person pushing a cart). This was because this distinction was deemed useful, and we didn't want to re-label for three groups if we just labeled two classes at the outset.

Results in Table \ref{tab:YOLO_table} show in general that the 3-class group with inter-class similarity 0.18 has lower accuracy across models than the 2-class group with similarity 0.15. For the 3-class option, one good choice that balances accuracy and size would be the sub-YOLO1 model (Y-1), whose accuracy is just $0.743 - 0.742 = 0.001$ less than the YOLO-nano model (Y-n), but whose size is 58\% ($1.1 / 1.9$) of YOLO-nano. When the \textit{person-walk} and \textit{person-cart} classes are merged into a single \textit{person} class, then sub-YOLO1 could be chosen with essentially the same accuracy as YOLOs (0.766 versus 0.764), but with 15\% (1.1 / 7.2) of the size.

% Pre
% \begin{table}[ht]
%   \caption{Results of detection of YOLOv5 small (Ys), nano (Yn) and the sub-YOLO models (Y-1, Y-2) on 3-class (person-walk, person-cart, robot) and 2-class (person, robot) cases respectively. The bracketed numbers below the model names are their model sizes. The accuracy is mAP@0.5.}
%   \label{tab:YOLO_table}
%   \centering
%   \footnotesize
%   \begin{tabular}{c c c c c c c c c}
%     \toprule
%     $N_{CL}$ & Simil. & Class & Ys & Yn & Y-1 & Y-2 \\
%      & & & \textcolor{red}{(7.2M)} & (1.9M) & \textcolor{red}{(1.1M)} & (0.16M) \\
%     \midrule
%     &  & p-walk & 0.728 & 0.727 & 0.725 & 0.665 \\
%     3 & 0.18 & p-cart & 0.690 & 0.674 & 0.681 & 0.576 \\
%     & & robot & 0.872 & 0.827 & 0.82 & 0.747 \\
%     \cmidrule{3-7}
%     & & mAP & \textcolor{red}{0.764} & 0.743 & 0.742 & 0.663 \\
%     \cmidrule{1-7}
%     &  & person & 0.752 & 0.732 & 0.719 & 0.691 \\
%     2 & 0.15 & robot & 0.875 & 0.827 & 0.814 & 0.755 \\
%      \cmidrule{3-7}
%      & & mAP & \textbf{0.813} & \textbf{0.779} & \textcolor{red}{\textbf{0.766}} & \textbf{0.723} \\
%      & & change & 6.4\% & 4.9\% & 3.2\% & 9.1\% \\
%     \bottomrule
%   \end{tabular}
% \end{table}

\begin{table}[ht]
  \caption{Results of detection of YOLOv5 small (Ys), nano (Yn) and the sub-YOLO models (Y-1, Y-2) on 3-class (person-walk, person-cart, robot) and 2-class (person, robot) cases respectively. The bracketed numbers below the model names are their model sizes. The accuracy is mAP@0.5.}
  \label{tab:YOLO_table}
  \centering
  \footnotesize
  % \begin{tabular}{p{8pt}p{10pt}p{30pt}p{20pt}p{20pt}p{20pt}p{20pt}p{20pt}p{20pt}p{20pt}}
  \begin{tabular}{cccccccccc}
    \toprule
    $N_{CL}$ & Simil. & Class & Ys & Yn & Y-1 & Y-2 \\
     & & & \textcolor{red}{(7.2M)} & (1.9M) & \textcolor{red}{(1.1M)} & (0.16M) \\
    \midrule
    &  & p-walk & 0.728 & 0.727 & 0.725 & 0.665 \\
    3 & 0.18 & p-cart & 0.690 & 0.674 & 0.681 & 0.576 \\
    & & robot & 0.872 & 0.827 & 0.82 & 0.747 \\
    \cmidrule{3-7}
    & & mAP & \textcolor{red}{0.764} & 0.743 & 0.742 & 0.663 \\
    \cmidrule{1-7}
    &  & person & 0.752 & 0.732 & 0.719 & 0.691 \\
    2 & 0.15 & robot & 0.875 & 0.827 & 0.814 & 0.755 \\
     \cmidrule{3-7}
     & & mAP & \textbf{0.813} & \textbf{0.779} & \textcolor{red}{\textbf{0.766}} & \textbf{0.723} \\
     & & change & 6.4\% & 4.9\% & 3.2\% & 9.1\% \\
    \bottomrule
  \end{tabular}
\end{table}

\subsection{Embedded Systems Examples}
\label{sec:subModelSelect}
We examine two aspects of efficiency, low power and small memory size, which in turn affect running and system cost for three examples of mobile, embedded system platforms. Some of the most energy-critical \citep{al2025energy, kinnas2025reducing, zhang2025high} types of systems are those running on battery power, so our experiments include mobile robots, drones, and IoT devices \citep{naveen2025optimized}. The scope of this paper is not to perform a comprehensive study of a range of these platforms, but to give an idea of how the aspects of this paper apply in practice.

For this experiment, we measured the relative energy use for a mobile robot, drone, and IoT device, where ``relative'' refers to the amount of energy needed for different levels of neural network processing. For a mobile robot and drone, the non-neural-network energy use is predominantly for wheel and propeller locomotion respectively. The mobile robot in our experiment is a small (17kg) Clearpath Jackal with LTE and WiFi, Velodyne LiDAR, video camera, and i-5 2.7GHz CPU. The drone is a small (17g) ModalAI VOXL-2 with LTE and WiFi, video camera, and Qualcomm QRB5165 AI processor. For consistency across comparisons, we fix the processor, model, and application to be the same, that is the TensorFlow-Lite (TFLite), 2-class model to detect horizontal and vertical shelf components. (The goal of this object detection application is for camera-based inventory monitoring in a warehouse). For each platform, we measured the average watts used with and without the detection algorithm running. In Table \ref{tab:DeviceEnergy}, one can see that the power used for locomotion of the robot and drone vastly overwhelms that used by the algorithm, and furthermore, there is $<1\%$ savings for using the smaller Y-4 model versus YOLO-nano. There is, however, a substantial reduction of required memory, 64\% from YOLO-nano to Y-4, and time reduction (average inference time) of 31\%.

Although our experiment shows memory and time savings but little power savings for the model-size reduction relative to the energy for locomotion, there is more to learn by a broader comparison of more models, devices, and applications. For instance, in \citep{ogorman2024pathplan}, we found a substantial energy savings (plus memory and time) for a comparison using a larger GPU device (NVIDIA GeForce GTX 1060) on a different application (fixed-camera robot path planning). 

% Pre
% \begin{table}[ht]
%   \caption{Robot, drone, and IoT device efficiencies.}
%   \label{tab:DeviceEnergy}
%   \centering
%   \footnotesize
%   \begin{tabular}{c | c |c c c | c}
%     Model & Memory &  & Power [w] & & Time \\
%      & [MB] & Robot & Drone & IoT & [ms]\\
%     \toprule
%     Baseline & & 97.5 & 52.1 & 2.5\\
%     Y-nano & 6.7 & 99.7 & 54.2 & 4.66 & 51.2 \\
%      Y-2 & 4.3 & 99.4 & 54.0 & 4.44 & 45.2 \\
%      Y-4 & 2.4 & 99.3 & 53.9 & 4.37 & 35.2 \\
%   \end{tabular}
% \end{table}

\begin{table}[ht]
  \caption{Robot, drone, and IoT device efficiencies.}
  \label{tab:DeviceEnergy}
  \centering
  \footnotesize
  \begin{tabular}{c | c |p{20pt} p{35pt} p{20pt} | p{10pt}}
  \toprule
    Model & Memory &  & Power [w] & & Time \\
     & [MB] & Robot & Drone & IoT & [ms]\\
    \midrule
    Baseline & - & 97.5 & 52.1 & 2.5 & - \\
    Y-nano & 6.7 & 99.7 & 54.2 & 4.66 & 51.2 \\
     Y-2 & 4.3 & 99.4 & 54.0 & 4.44 & 45.2 \\
     Y-4 & 2.4 & 99.3 & 53.9 & 4.37 & 35.2 \\
     \bottomrule
  \end{tabular}
\end{table}

%-------------------------------------------------------------------------

\section{Conclusions}
\label{sec:Conclusions}

% There are 
% \bo{We summarize} three high level messages \bo{distilled from our investigation} in this paper. 
Through this work, we elucidate three high-level messages that support our findings. i) Few class ($<10$) image classification has distinctive properties that are different from classification of a larger number of classes. ii) This few-class distinctiveness is one aspect of data-side properties, which we believe to be under-explored in the literature. Conversely model-side design, which is supported by benchmark testing is largely focused upon large datasets, which as we imply from (i), do not give full insight into few-class performance. iii) There are many few-class applications in practice, so methodologies that can help the practitioner with efficient model selection can offer the following benefits: reduce time and effort in model selection, select models with better performance (some combination of accuracy, size, speed, power usage, and cost), and to expand on power usage, attain better energy sustainability of models whose applications may be deployed in the thousands or millions.

Four technical contributions support the messages of this paper, all focused and distinctive to the few-class regime. i) We compared classification difficulty measures and identified weightings of intra- and inter-class similarity that best correlate with accuracy. ii) We investigated the contributions (i.e., highest correlated weightings) of intra- and inter-class similarities and made two observations. Intra-class similarity contributes more highly than inter-class similarity for few-classes (and for the models and datasets tested), and the effect of inter-class similarity on accuracy rises toward that of intra-class similarity as the number of classes increases. iii) The dataset difficulty metric can be used to predict relative classification performance of different class groupings or datasets. Use of this metric is much faster than conventional training and testing for comparing different class groupings. iv) We have proposed using ``sub-models'', that is models smaller than the lowest published model in scaled model families such as EfficientNet and YOLO. 

We have endeavored to provide extensive experimentation of models and datasets to support our findings on few-class distinctiveness and utility. For the investigation of difficulty metrics in Section \ref{sec:difficMeasure}, we trained and tested 6000 trained models in total on 3 architectures (MobileNet-V2, ResNet-50, and ViT, 5 sizes of number of classes from CIFAR-100 and ImageNet-200, and 200 random selection of classes per subsets. For the investigation of few-class properties in Section \ref{sec:nCl}, we trained and tested 250 models for 5 ResNet sizes from ResNet-152 to ResNet-18, 10 sizes of number of classes from ImageNet-1000 from 1000 to 2 classes, and 5 subset selections per number of classes. In Section \ref{sec:Exp_intersim}, we trained and tested on 18 class groupings for 3 models, EfficientNet-B0, VGG-19, and MobileNet-V2; plus 45 pairwise class groupings of CIFAR-100. And we trained and tested 9 levels of YOLO for 3 class groupings on a total of 89 subsets of the COCO dataset in Section \ref{sec:InPractice}.

As far as limitations of this work, we suggest a few here. i) Although we trained and tested extensively on over 6000 models, we have been careful to include the caveat that these results apply ``to the datasets tested''. We found consistent results for the CIFAR-100 and ImageNet-1000 datasets of Section \ref{sec:difficMeasure} in which similarities from many class groupings were extracted. However, testing for some more focused datasets such as Food-101 \citep{food101_2014} and Textures-47 \citep{textures47_2014} might have different balances of intra- and inter-class similarities, which may lead to different weightings of the difficulty metric in equation \ref{eqn:DifficF}. ii) As is evident in Fig.s \ref{fig:RVsNc6plots} and \ref{fig:ResNet_fewClass} few-class properties appear to converge with many-classes as the number of classes increases, but we didn't extend the comparison of models and datasets above $N_{CL} = 10$ (due to time- and energy-use considerations of the larger sub-sets and explosion of combinations for larger class groupings). iii) Our testing of one robot, one drone, and one IoT device platform in Section \ref{sec:applSpec} and \ref{sec:subModelSelect} was intended only to offer a sampling of how few-class properties are important in practice. iv) Finally, we did not cover the data-side property of scale-resolution for reasons described in Section \ref{sec:Exp_color}. We suggest that topics in this paragraph should be undertaken in future work.

% \section{Acknowledgement}
% \label{sec:ack}
% This research has been supported in part by the National Science Foundation (NSF) under Grant No. CNS-2055520.

\newpage
%% The Appendices part is started with the command \appendix;
%% appendix sections are then done as normal sections
\appendix

\section{Data Availability Statement}
\label{app1}

% Error >>
% \subsection{Datasets}
Dataset information is presented in Table \ref{tab:ten_ds_info}, including the dataset name, abbreviation, reference, the link to the dataset and path in \FCA~.

% \clearpage
\begin{table*}[h]
    \centering
    \tiny
    % \footnotesize
    % \begin{tabular}{p{2.5em}|ccccccccccc}
    \begin{tabular}{p{4.5em}p{2.2em}p{6em}p{26em}p{13.5em}}
        \toprule
        \textbf{Dataset Name} & \textbf{Abbrev.} & \textbf{Ref.} & \textbf{Homepage} & \textbf{Path in FCA} \\
        \midrule
        \midrule
        CIFAR-10 & CF10 & \citep{krizhevsky2009learning} & \href{cs.toronto.edu/~kriz/cifar.html}{cs.toronto.edu/\~kriz/cifar.html} & - \\
        Common Objects in Context & COCO & \citep{lin2014microsoft} & \href{cocodataset.org}{cocodataset.org} & - \\

        \midrule
        \midrule
        Caltech 101 & CT101 & \citep{li_andreeto_ranzato_perona_2022}~ & \href{data.caltech.edu/records/mzrjq-6wc02}{data.caltech.edu/records/mzrjq-6wc02} & \href{tools/ncls/datasets/caltech101.py}{tools/ncls/datasets/caltech101.py} \\

        \midrule
        Caltech 256 & CT256 & \citep{griffin_holub_perona_2022}~ & \href{data.caltech.edu/records/nyy15-4j048}{data.caltech.edu/records/nyy15-4j048} & \href{tools/ncls/datasets/caltech256.py}{tools/ncls/datasets/caltech256.py} \\

        \midrule
        CIFAR-100 & CF100 & \citep{krizhevsky2009learning}~ & \href{cs.toronto.edu/~kriz/cifar.html}{cs.toronto.edu/~kriz/cifar.html} & \href{tools/ncls/datasets/cifar100.py}{tools/ncls/datasets/cifar100.py} \\
        & & & \href{github.com/knjcode/cifar2png}{github.com/knjcode/cifar2png} & \\

        \midrule
        Caltech-UCSD & CB200 & \citep{wah_branson_welinder_perona_belongie_2011}~ & \href{vision.caltech.edu/visipedia/CUB-200-2011.html}{vision.caltech.edu/visipedia/CUB-200-2011.html} & \href{tools/ncls/datasets/cub200.py}{tools/ncls/datasets/cub200.py} \\
        Birds-200-2011 & & & \href{data.caltech.edu/records/65de6-vp158/files/CUB_200_2011.tgz}{data.caltech.edu/records/65de6-vp158/files/CUB\_200\_2011.tgz} & \\

        \midrule
        Food 101 & FD101 & \citep{bossard14}~ & \href{vision.ee.ethz.ch/datasets_extra/food-101/}{vision.ee.ethz.ch/datasets\_extra/food-101/} & \href{tools/ncls/datasets/food101.py}{tools/ncls/datasets/food101.py} \\
        & & & \href{huggingface.co/datasets/food101}{huggingface.co/datasets/food101} & \\
        
        \midrule
        German Traffic Sign & GT43 & \citep{Stallkamp2012}~ & \href{benchmark.ini.rub.de/}{benchmark.ini.rub.de/} & \href{tools/ncls/datasets/gtsrb43.py}{tools/ncls/datasets/gtsrb43.py} \\
        % Recognition Benchmark & \\

        \midrule
        ImageNet & IN1K & \citep{deng2009imagenet}~ & \href{image-net.org/challenges/LSVRC/2012/index.php}{image-net.org/challenges/LSVRC/2012/index.php} & * \\

        \midrule
        Indoor Scene Recognition & ID67 & \citep{quattoni2009recognizing}~ & \href{web.mit.edu/torralba/www/indoor.html}{web.mit.edu/torralba/www/indoor.html} & \href{tools/ncls/datasets/indoor67.py}{tools/ncls/datasets/indoor67.py} \\
         % & \\

        \midrule
        Quickdraw & QD345 & \citep{ha2017neural}~ & \href{github.com/googlecreativelab/quickdraw-dataset}{github.com/googlecreativelab/quickdraw-dataset} & \href{tools/ncls/datasets/quickdraw345.py}{tools/ncls/datasets/quickdraw345.py} \\
        & & & \href{tensorflow.org/datasets/community_catalog/huggingface/quickdraw}{tensorflow.org/datasets/community\_catalog/huggingface/quickdraw} & \\

        \midrule
        Describable Textures Dataset & TT47 & \citep{cimpoi14describing}~ & \href{robots.ox.ac.uk/~vgg/data/dtd/index.html}{robots.ox.ac.uk/~vgg/data/dtd/index.html} & \href{tools/ncls/datasets/textures47.py}{tools/ncls/datasets/textures47.py} \\
         % & \\
        \bottomrule
    \end{tabular}
    \caption{Dataset information. * Note that ImageNet dataset format is used as the reference for other datasets. Therefore, the Path in FCA is not required for ImageNet.}
    \label{tab:ten_ds_info}
\end{table*}
% Error <<
\clearpage

\noindent \textbf{License.} We have searched available online resources and list the license of each dataset in Table A.9. For licenses not found in the datasets or websites denoted as ``*'', we assume they are non-commercial research use only.

\begin{comment}
\noindent \textbf{Train/val splits.} 
The dataset format follows the convention of ImageNet:
\begin{verbatim}
imagenet1k/
    ├── meta
    │   ├── train.txt
    │   └── val.txt
    ├── train
    │   ├── <IMAGE_ID>.jpeg
    │   └── ...
    └── val
        ├── <IMAGE_ID>.jpeg
        └── ...
\end{verbatim}
where a .txt file stores a pair of image id and and class number in each row in the following format:

\begin{verbatim}
<IMAGE_ID>.jpeg <CLASS_NUM>
\end{verbatim}

% \begin{table*}[h]
% \footnotesize
%     \begin{center}
%     % \begin{tabular}{p{2.5em}|ccccccccccc}
%     \begin{tabular}{p{6em}p{8em}|p{6em}p{8em}}
%         \toprule
%         \textbf{Dataset} & \textbf{License} & \textbf{Dataset} & \textbf{License} \\

%         \midrule
%         CT101 & CC BY 4.0 & CT256 & CC BY 4.0 \\
%         CF10, CF100 & MIT & CB200 & CC BY 4.0 \\
%         FD101 & CC BY-SA 4.0 & GT43 & GPLv2 \\
%         IN1K & * & ID67 & * \\
%         QD345 & CC BY 4.0 & TT47 & * \\
%         \bottomrule
%     \end{tabular}
%             \caption{Licenses of ten datasets.}
%     \end{center}
%     \label{tab:license}
% \end{table*}
\end{comment}

% \clearpage
\begin{table}[h]
\footnotesize
    \begin{center}
    % \begin{tabular}{p{2.5em}|ccccccccccc}
    \begin{tabular}{p{6em}p{8em}}
        \toprule
        \textbf{Dataset} & \textbf{License} \\
        \midrule
        CT101 & CC BY 4.0 \\
        CT256 & CC BY 4.0 \\
        CF10, CF100 & MIT \\
        CB200 & CC BY 4.0 \\
        FD101 & CC BY-SA 4.0 \\
        GT43 & GPLv2 \\
        IN1K & * \\
        ID67 & * \\
        QD345 & CC BY 4.0 \\
        TT47 & * \\
        \bottomrule
    \end{tabular}
            \caption{Licenses of ten datasets.}
    \end{center}
    \label{tab:license}
\end{table}
% \clearpage

We follow the same train/val splits when the original dataset has already provided. If the dataset does not have explicit splits, we first assign image IDs to all images, starting from 0. We then select $4/5$ of the images as the training set and place the rest in the validation set. Specifially, if an image's ID satisfies the condition $ID$ $\%$ $5 == 0$, it is moved to the validation set; otherwise, it is assigned as a training sample.

Due to confidentiality and proprietary business considerations, datasets in Sec. \ref{sec:applSpec} are not publicly released.

\clearpage

% =========================================================

% \acks{All acknowledgements go at the end of the paper before appendices and references.
% Moreover, you are required to declare funding (financial activities supporting the
% submitted work) and competing interests (related financial activities outside the submitted work).
% More information about this disclosure can be found on the JMLR website.}
\acks{This research has been supported in part by the National Science Foundation (NSF) under Grant No. CNS-2055520.}

% Manual newpage inserted to improve layout of sample file - not
% needed in general before appendices/bibliography.

% \newpage

\vskip 0.2in
\bibliography{main}

\end{document}